\documentclass[conference]{IEEEtran}
\IEEEoverridecommandlockouts
\usepackage{cite}
\usepackage{amsmath,amssymb,amsfonts}
\usepackage{algorithmic}
\usepackage{graphicx}
\usepackage{textcomp}
\usepackage{xcolor}

\newcommand{\x}{\bm{x}}
\newcommand{\xg}{\bm{x}_\gamma}

\newcommand{\eps}{\bm{\epsilon}}

\newcommand{\E}{E_\theta} 
\newcommand{\m}{m_\theta} 
\newcommand{\R}{\mathbb{R}}
\newcommand{\Ex}{\mathbb{E}}
\newcommand{\norm}[1]{\lVert #1 \rVert}
\newcommand{\ip}[2]{\langle #1, #2 \rangle}

\newcommand{\eqme}{EqM-E}
\newcommand{\ours}{CEqM}

\usepackage{pifont}
\newcommand{\cmark}{\ding{51}}
\newcommand{\xmark}{\ding{55}}

\usepackage{bm}
\usepackage{graphicx}
\usepackage{booktabs}
\newcommand{\tabstretch}{1}
\usepackage{float}
\usepackage{tabularray}
\usepackage{threeparttable}
\usepackage{amsmath,amssymb,amsthm,mathtools}

\theoremstyle{plain}
\newtheorem{proposition}{Proposition}
\newtheorem{lemma}{Lemma}
\theoremstyle{definition}
\newtheorem{example}{Example}
\usepackage{capt-of}

\usepackage[hidelinks]{hyperref}
\usepackage[capitalize]{cleveref}
\crefname{section}{Sec.}{Secs.}
\Crefname{section}{Section}{Sections}
\Crefname{table}{Table}{Tables}
\crefname{table}{Table}{Tables}
\crefname{figure}{Fig.}{Figs.}
\Crefname{figure}{Fig.}{Figs.}

\usepackage{xspace}
\makeatletter
\DeclareRobustCommand\onedot{\futurelet\@let@token\@onedot}
\def\@onedot{\ifx\@let@token.\else.\null\fi\xspace}
\def\etal{\emph{et al}\onedot}
\def\vs{\emph{vs}\onedot}
\def\eg{\emph{e.g}\onedot}
\def\ie{\emph{i.e}\onedot}
\makeatother

\def\BibTeX{{\rm B\kern-.05em{\sc i\kern-.025em b}\kern-.08em
    T\kern-.1667em\lower.7ex\hbox{E}\kern-.125emX}}
\begin{document}

\title{Give Me a Brake! Bridging Implicit and Explicit Energy Landscapes with Conformal Equilibrium Matching}

\author{\IEEEauthorblockN{Lorenzo S. Querol}
\IEEEauthorblockA{\textit{Graduate School of Information Science and Technology} \\
\textit{The University of Osaka} \\
Osaka, Japan \\
lorenzoquerol@is.ids.osaka-u.ac.jp}
\and
\IEEEauthorblockN{Hideaki Hayashi}
\IEEEauthorblockA{\textit{D3 Center} \\
\textit{The University of Osaka}\\
Osaka, Japan \\
hayashi@ids.osaka-u.ac.jp}
\and
\IEEEauthorblockN{Hajime Nagahara}
\IEEEauthorblockA{\textit{D3 Center} \\
\textit{The University of Osaka}\\
Osaka, Japan \\
nagahara@ids.osaka-u.ac.jp}
}

\IEEEaftertitletext{%
  \centering
  \includegraphics[width=0.9\textwidth]{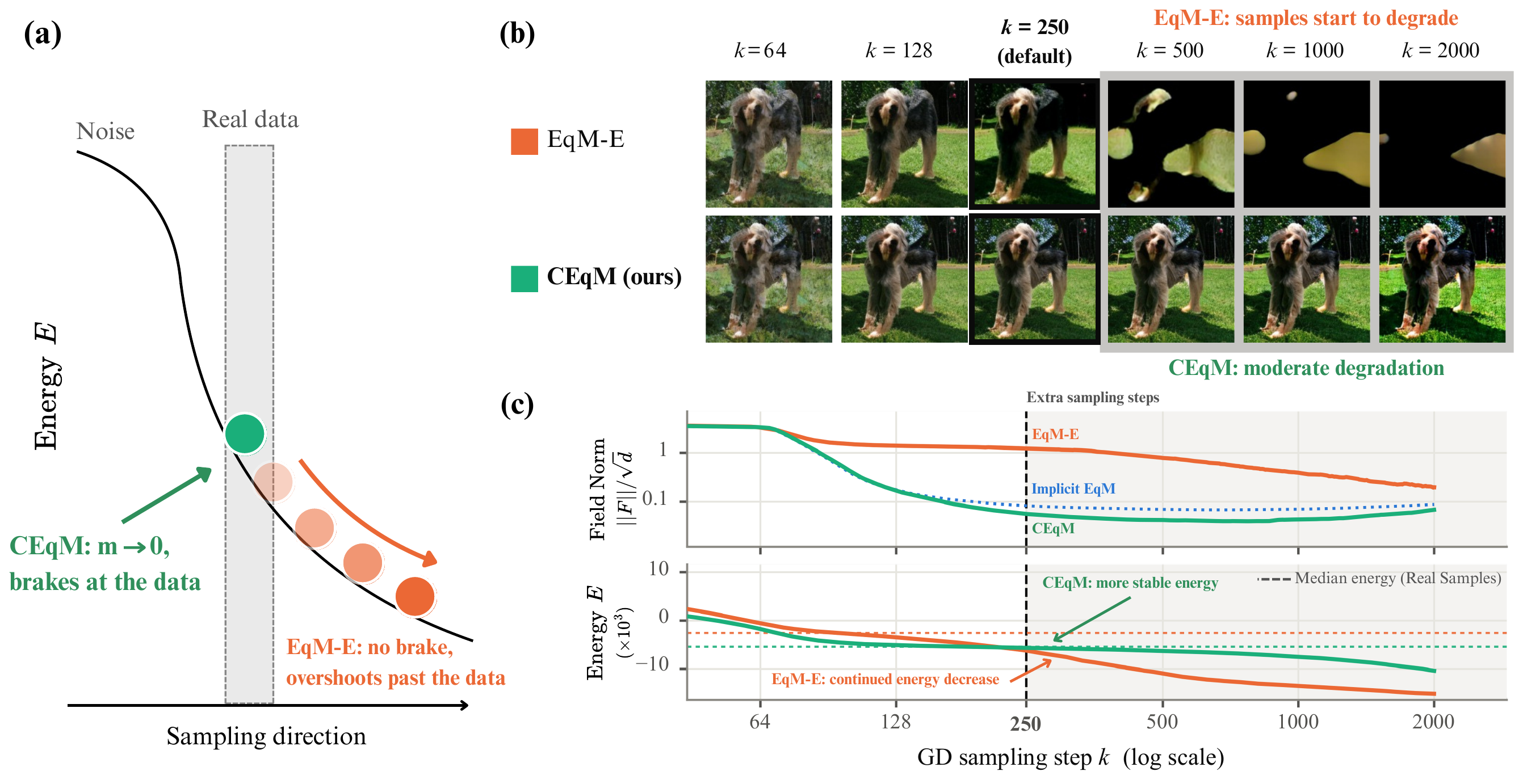}
  \captionof{figure}{\textbf{Give me a brake!} Equilibrium matching (EqM) samples by gradient descent, which should slow down as it reaches the data. \textbf{(a)} With an explicit energy (\eqme, $F=\nabla E$), nothing slows the sampler and it rolls past the data. Our conformal EqM (\ours{}) scales the gradient by a learned gain, $F=m\nabla E$, and $m\to0$ near the data acts as a brake. \textbf{(b)} Samples after $k$ steps. Sampling longer degrades \eqme's images, whereas \ours{} slows the degradation. \textbf{(c)}~Field norm and energy along the trajectory. The brake is temporary: \ours{} eventually rolls past the median real-data energy, suggesting real data are not minima of $E$.}
  \label{fig:teaser}
  \vspace{0.5\baselineskip}}
\maketitle

\begin{abstract}
Equilibrium matching (EqM) is a generative modeling framework that learns a single time-invariant field that vanishes at the data. It admits two parameterizations: implicit EqM, which directly predicts the field, and EqM with an explicit energy (\eqme{}), which parameterizes the field as an energy's gradient. \eqme{} provides the scalar energy $E$ that makes energy-based models useful, but performs worse than implicit EqM. We show that this gap stems from the symmetry of the fields representable by \eqme{}, which prevents it from fully expressing the field required by the EqM objective. Based on this insight, we propose conformal EqM (\ours{}), which factorizes the field as $m\nabla E$. The positive gain $m$ is predicted by adding a scalar-output head to the \eqme{} network, without changing the training objective or sampler. It can be interpreted as the reciprocal of the conformal metric on the energy's Riemannian gradient, acting as a \textbf{brake} on gradient descent during sampling, thereby restoring the vanishing field and step-size robustness of implicit EqM. At EqM-B/4, \ours{} improves FID from 60.2 (\eqme{}) and 59.0 (implicit EqM) to 46.1. It stays within 1.5 FID of implicit EqM at EqM-B/2 and on CIFAR-10, and within 2.7 on CIFAR-100, while keeping the benefits of an explicit energy, such as out-of-distribution detection and energy composition.
\end{abstract}


\section{Introduction} \label{sec:intro}
Energy-based models (EBMs) learn a single scalar energy function that assigns low values to likely input configurations and high values to unlikely ones, thereby creating an energy landscape that models a data distribution~\cite{lecun2006tutorial}. Learning this energy comes with several advantages, such as sample generation~\cite{du2019implicit}, composition~\cite{du2020compositional,liu2022compositional,du2023reduce}, and input scoring~\cite{grathwohl2020jem, liu2020energyood}. Well-known methods to learn this energy function include contrastive divergence~\cite{hinton2002cd,du2021improved}, short-run Markov chain Monte Carlo~(MCMC)~\cite{nijkamp2019shortrun}, and diffusion recovery likelihood~\cite{gao2021drl, zhu2024cdrl}. Despite these advances, training EBMs has remained a challenge due to the intractable normalizing constant, and MCMC-based methods are known to be costly and unstable. These difficulties have limited their scalability and sample quality compared with diffusion- and flow-based models~\cite{ho2020denoising,song2021sde,lipman2023flow,ma2024sit}.

Much of the recent progress in EBMs departs from established methods and adopts a flow-based training approach. Equilibrium matching (EqM)~\cite{wang2025eqm} offers a regression-based approach to learning generative equilibrium dynamics. Instead of drawing negative samples during training, it matches a vector field to a closed-form target defined along interpolations between data and noise. Unlike diffusion and flow models, EqM learns a single time-invariant field without observing the noise level. Its target decays to zero at the data, making real samples stationary points under ideal training. Sampling also supports an optimization-based procedure, with flexible step sizes, optimizers, and stopping criteria.  With this formulation, EqM combines simple training without MCMC with strong generation quality and the flexibility of equilibrium-based sampling.

EqM offers two parameterizations with distinct advantages and limitations. Implicit EqM directly outputs the vector field, achieving strong generation quality without explicitly modeling a scalar energy. Consequently, it does not provide directly evaluable energy values. EqM with an explicit energy (\eqme{}) instead models a scalar function and uses its input gradient as the field. This makes the energy directly accessible for applications such as out-of-distribution (OOD) detection and energy composition. However, this additional capability comes with a substantial gap in generation quality relative to implicit EqM as reported in~\cite{wang2025eqm}. Retaining an explicit energy while closing this gap is the central goal of this work.

We show that this performance gap stems from the symmetry of the fields representable by \eqme{}, which prevents it from adequately representing the field required by the EqM objective. Because its field is the gradient of a scalar function, its Jacobian must be symmetric; it cannot represent rotational components. The field required by EqM need not satisfy this constraint. Although the optimal target at each fixed noise level is a gradient, hiding that level requires averaging these targets with weights that depend on the input. Such an average can have an asymmetric Jacobian and therefore need not be a gradient. Implicit EqM directly predicts the field and is free from this symmetry constraint, whereas \eqme{} incurs a representation error wherever this asymmetry occurs (Sec.~\ref{sec:diagnosis}).

Motivated by this insight, we propose conformal EqM (\ours{}), which decomposes the vector field as $m\nabla E$. The energy gradient $\nabla E$ supplies the direction, while a learned positive gain $m$ controls the magnitude. Because the gain varies with the input, the resulting field can have an asymmetric Jacobian, relaxing the symmetry constraint of \eqme{} while retaining an explicit energy. Positivity ensures that the continuous sampling flow descends this energy. The gain can be predicted by adding a scalar-output head to the EqM-E architecture, while leaving the EqM training objective and sampler unchanged (Sec.~\ref{sec:method}).

The positive gain also gives \ours{} a geometric interpretation and a practical braking mechanism. Rescaling the Euclidean metric by $1/m$ defines a conformal metric under which $m\nabla E$ is the energy's Riemannian gradient, motivating the name conformal EqM. During sampling, the gain acts as a brake on gradient descent: it shrinks as samples approach the data, slowing the updates even when the energy gradient remains nonzero. \ours{} recovers two desirable properties of implicit EqM in practice: the field decays toward zero at the data (Figs.~\ref{fig:teaser} and \ref{fig:field_norms}), and sampling is robust to the choice of step size (Appendix~\ref{sec:supp_eta}).

\noindent In summary, we contribute the following: 
\begin{enumerate}
    \item We identify the symmetry constraint of explicit energy gradients as a limitation of \eqme{} and show how hiding the noise level can make the optimal EqM field violate this constraint.
    \item We propose \ours{}, which learns an explicit energy and a positive gain to relax this constraint, admits a conformal Riemannian interpretation, and preserves energy descent along the continuous sampling flow without changing the training objective or sampler.
\end{enumerate}

\section{Related Work} \label{sec:related_work}

\subsection{Noise-Unconditional Generative Models} 
Diffusion and flow models~\cite{sohl2015deep,song2019generative,ho2020denoising,song2021denoising,song2021sde,karras2022edm,lipman2023flow,liu2023rectified,albergo2023interpolants, peebles2023dit,ma2024sit} learn a field that is conditioned on different noise levels. Sun~\etal~\cite{sun2025noise} showed that removing this conditioning degrades most models only gracefully, thereby linking them to blind denoising~\cite{mohan2020biasfree}, in which the network infers the noise level from its input. Kadkhodaie and Simoncelli~\cite{kadkhodaie2021stochastic} sample from a blind denoiser with step sizes set by the noise level estimated from its residual. Sahraee-Ardakan~\etal~\cite{sahraee2026geometry} analyze the optimal noise-unconditional field and find that, up to correction terms, it is the gradient of the marginal energy $-\log p(\x)$ multiplied by a posterior-averaged gain, a conformal factor that turns the singular potential well at the data into stable attractors. For blind Gaussian denoising, we note that the correction is itself a gradient, so the optimal field is exactly a gain times the gradient of an energy. In this work, we show what is lost when this gain is absorbed into the energy~(Sec.~\ref{sec:diagnosis}). Thus, by decomposing the field into a positive scalar gain and energy, we restore the desirable properties that a noise-unconditional field offers.

\begin{figure*}[t]
    \centering
    \includegraphics[width=\textwidth]{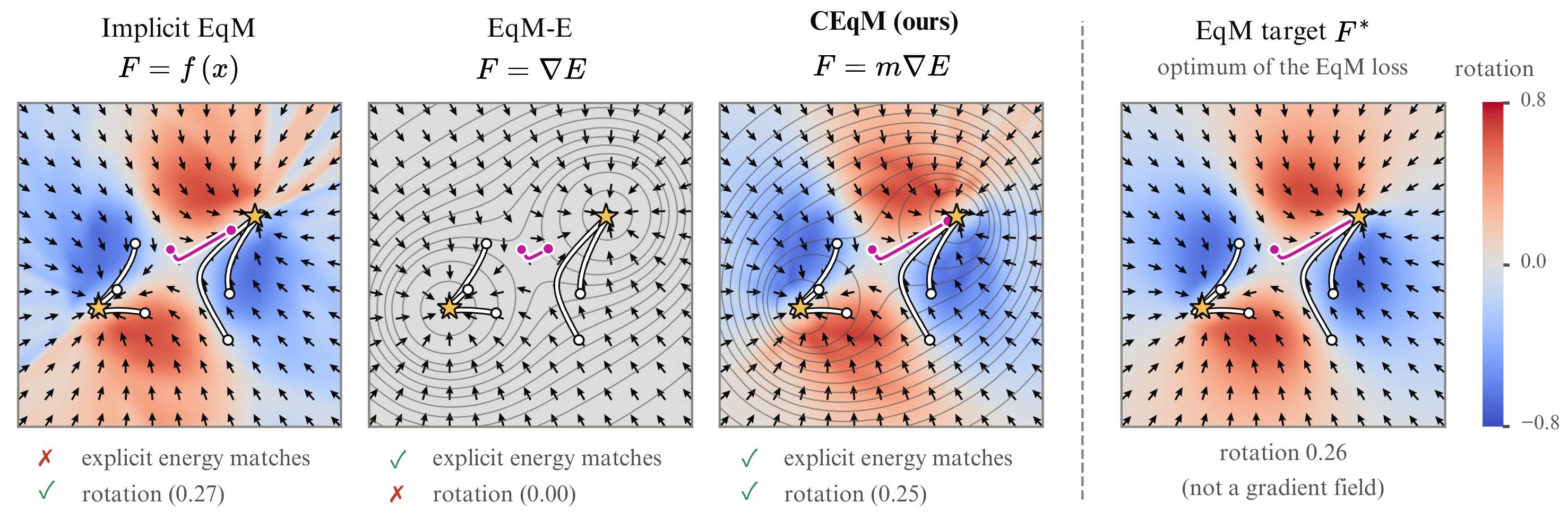}
    \caption{\textbf{Learned Vector Fields on 2D Examples.} We compare the learned vector fields and sample trajectories along different EqM variants. Arrows show the field $F$, while color shows its curl, which is zero for a gradient field. Numbers give their mean absolute value over the grid. White paths are gradient-descent trajectories. We highlight interesting trajectories in magenta that show where samples stop before stationary points (stars).}
  \label{fig:toy_example}
\end{figure*}

\subsection{Energy-Based Models} 
EBMs~\cite{lecun2006tutorial} define an unnormalized density through a scalar energy. The classical way of training them by contrastive divergence with MCMC~\cite{hinton2002cd,xie2016convnet,du2019implicit,nijkamp2019shortrun,grathwohl2020jem,du2021improved} is known to be costly and unstable. Diffusion recovery likelihood~\cite{gao2021drl,zhu2024cdrl} narrowed the quality gap by learning a sequence of noise-conditional EBMs. Recent methods instead learn a single time-independent energy by regressing its gradient onto a transport field. Energy Matching~\cite{balcerak2025energy} combines an optimal-transport flow far from the data with contrastive refinement near it, Projected Energy Matching~\cite{barco2026projected} fits an energy to an unconstrained velocity teacher, and Wang~\etal~\cite{wang2026lyapunov} analyze such energies with Lyapunov theory. Rodriguez~\etal~\cite{rodriguez2026generative} train an Energy Transformer whose forward pass descends a global energy function using the EqM objective. The explicit energy variant of EqM~\cite{wang2025eqm} regresses the gradient of a scalar energy directly onto the EqM target, which is the central focus of this work.

\subsection{Conservativity of Learned Fields}
Whether a learned field should be constrained to be a gradient has been debated in the context of diffusion models. Salimans and Ho~\cite{salimans2021energy} found that energy-parameterized score models match unconstrained ones when the architectures are comparable, and Du~\etal~\cite{du2023reduce} adopt the energy parameterization for composition and Metropolis-corrected sampling. Chao~\etal~\cite{chao2023qcsbm} showed that unconstrained score models are measurably non-conservative and proposed a regularizer, and Horvat and Pfister~\cite{horvat2024gauge} showed that exact sampling does not require conservativity. For noise-conditional models, the constraint has no cost at the optimum because the target at each level is itself a gradient. However, for time-invariant models, the target is a single field that does not need to be a gradient, and fitting $\nabla E$ to it in $L_2$ returns its projection onto the conservative field~\cite{barco2026projected}. While prior work has analyzed noise-unconditional fields and learned energies from transport fields, we identify the symmetry constraint of explicit energy gradients as a limitation in EqM. To relax this constraint, we learn a positive scalar gain alongside the energy, allowing the resulting field to be non-conservative while retaining an explicit energy. This approach requires no change to the EqM training objective or sampler.

\section{Preliminaries}\label{sec:preliminaries}
\subsection{Equilibrium Matching} 
EqM \cite{wang2025eqm} distinguishes itself from conventional flow-matching approaches by learning a time-invariant vector field where the gradient vanishes on the data manifold, and ground-truth samples are treated as stationary points. Given a corrupted intermediate sample $\xg=\gamma \bm{x}+(1-\gamma)\eps$, where $\bm{x}$ is a training image and $\eps \sim \mathcal{N}(\bm{0}, \bm{I})$, the loss objective is defined as:
\begin{equation}
    \mathcal{L}_{\text{EqM}}= \mathbb{E}_{\x,\eps,\gamma} \left[\frac{1}{2}\| \bm{F}(\xg) - \bm{v}(\xg) \|^2\right]
    \label{eq:eqm_objective}
\end{equation}
where $\bm{F}$ is the output vector field, $\bm{v}(\xg)=c(\gamma)(\eps-\bm{x})$ is the target vector field, and $c(\gamma)$ controls the gradient magnitude along the interpolation path, satisfying $c(1)=0$. The network never observes $\gamma$. This constraint ensures that the gradient vanishes at real data points under ideal training, so that true samples correspond to stationary points of the learned landscape. The notation and the regularity assumptions used in our analysis of this objective, together with the precise sense in which real samples are stationary points of its minimizer, are given in Appendices~\ref{sec:supp_notation} and~\ref{sec:supp_optimal}.

\subsection{Implicit and Explicit Energies}
EqM admits two parameterizations: implicit EqM and \eqme{}~\cite{wang2025eqm}. In implicit EqM, the network directly outputs the vector field $\bm{F}$ without explicitly modeling a scalar energy. This formulation achieves strong generation quality but does not provide directly evaluable energy values. In contrast, \eqme{} models a scalar energy function $\E$, where $\theta$ denotes the network parameters, and sets $\bm{F}(\xg)=\nabla_{\xg}\E(\xg)$ to compute the same loss in Eq.~\eqref{eq:eqm_objective}, taking the gradient with respect to the corrupted input $\xg$. Through this parameterization, EqM learns an energy landscape in which real samples correspond to lower energies and noisy samples to higher energies. The ability to explicitly evaluate $\E$ enables applications such as OOD detection, but the reported generation quality of \eqme{} is lower than that of implicit EqM~\cite{wang2025eqm}.

\section{Why Explicit Equilibrium Matching Falls Short: The Equilibrium Field Is Not a Gradient} \label{sec:diagnosis}
The reason for the quality gap between \eqme~and implicit EqM can be explained in one sentence: \eqme~can only represent fields that are the gradient of a scalar function, but the field that the EqM loss asks the network to produce is \emph{not} a gradient. This is a modeling bias rather than an optimization or capacity issue, so more training does not remove it. We keep the mathematics in this section to a minimum; precise statements and proofs can be found in Appendices~\ref{sec:supp_symmetry}--\ref{sec:supp_structure}.

\subsection{What an Explicit Energy Can and Cannot Represent}
\Cref{fig:toy_example} shows learned vector fields on 2D examples where the target field has non-zero rotation (curl). The color coding shows that implicit EqM captures this rotation, while EqM-E, constrained to represent only gradient fields, approximates it with its curl-free component. This constraint can cause sampling trajectories in \eqme{} to stop before reaching stationary points. The figure thus provides empirical evidence for the theoretical limitation discussed below. As discussed later, our proposed \ours{} relaxes this constraint and can represent fields with a rotational component.

A field $\bm{F}$ is called \emph{conservative} if it is the gradient of some scalar function, $\bm{F}=\nabla E$. Conservative fields have a simple fingerprint: their Jacobian $\bm{J}_{\bm{F}}=\partial\bm{F}/\partial\bm{z}$ is a symmetric matrix at every point, because it is the Hessian of $E$. The converse also holds on $\R^D$, so the \emph{antisymmetric part} of the Jacobian,
\begin{equation}
  \bm{A}_{\bm{F}}(\bm{z})=\tfrac12\big(\bm{J}_{\bm{F}}(\bm{z})-\bm{J}_{\bm{F}}(\bm{z})^{\!\top}\big),
  \label{eq:antisym}
\end{equation}
measures exactly how far a field is from being a gradient. Fields with $\bm{A}_{\bm{F}} \neq 0$ have \emph{rotation}: they circulate around some points instead of only descending towards them, and no energy landscape can produce circulation. \eqme{} is by construction restricted to fields with $\bm{A}_{\bm{F}}\equiv0$. If the regression target of the loss~\eqref{eq:eqm_objective} has rotation, the best an explicit-energy model can do is to fit the non-rotating part of the target and discard the rest, no matter how large the network is or how long it is trained. The equivalence between gradient fields and symmetric Jacobians is proved in Appendix~\ref{sec:supp_symmetry}, which also makes precise in what sense $\bm{A}_{\bm{F}}$ measures the departure from a gradient field.

\subsection{What the Regression Objective Requests}
The network sees a corrupted sample $\bm{z}=\xg$ but not the interpolation factor $\gamma$ that produced it. For a given input $\bm{z}$, there are therefore many noise levels that could have produced it, and the loss is minimized by averaging the targets of all of them, weighted by how plausible each level is given $\bm{z}$:
\begin{equation}
\begin{aligned}
  \bm{F}^\star(\bm{z})&=\mathbb{E}\big[c(\gamma)(\eps-\x)\,\big|\,\xg=\bm{z}\big]\\
  &=\int_0^1 p(\gamma\mid\bm{z})\,c(\gamma)\,\underbrace{\mathbb{E}\big[\eps-\x\mid\bm{z},\gamma\big]}_{\text{per-level target}}\,d\gamma .
\end{aligned}
  \label{eq:fstar}
\end{equation}
Here, $p(\gamma\mid\bm{z})$ is the posterior over the noise level given the input. Moving the input changes which noise levels are plausible, thereby altering both the per-level targets and their weights. Differentiating the weights introduces an additional term that need not be symmetric. Thus, an average of gradient fields need not itself be a gradient. Appendix~\ref{sec:supp_optimal} derives Eq.~\eqref{eq:fstar}, shows by Tweedie's formula that every per-level target is a gradient field, gives the antisymmetric part of $\bm{F}^\star$ in closed form, and works out a two-dimensional Gaussian example in which $\bm{F}^\star$ is not a gradient.

\subsection{What This Means for Explicit EqM}
Where this asymmetry occurs, \eqme{} cannot exactly fit the optimal field. Under this condition, its best possible regression loss has an approximation gap that cannot be eliminated by increasing the network capacity or training length. Although this condition does not necessarily hold for every data distribution, the resulting gap is likely to affect the quality of generated images. The design question, therefore, is how to relax the symmetry constraint while retaining an explicit descent objective. The gap is proved in Appendix~\ref{sec:supp_gap}, and Appendix~\ref{sec:supp_structure} describes the structure of the asymmetry when the noise level can be inferred from the input.

\section{\ours{}: Conformal Equilibrium Matching} \label{sec:method}
Our proposed CEqM adds only a single input-dependent scalar to EqM-E. This relaxes the constraint of Jacobian symmetry while retaining the benefit of explicit energy estimation. Equivalently, it estimates the field by separating its magnitude from its direction.

\begin{figure}[t]
  \centering
  \includegraphics[width=\linewidth]{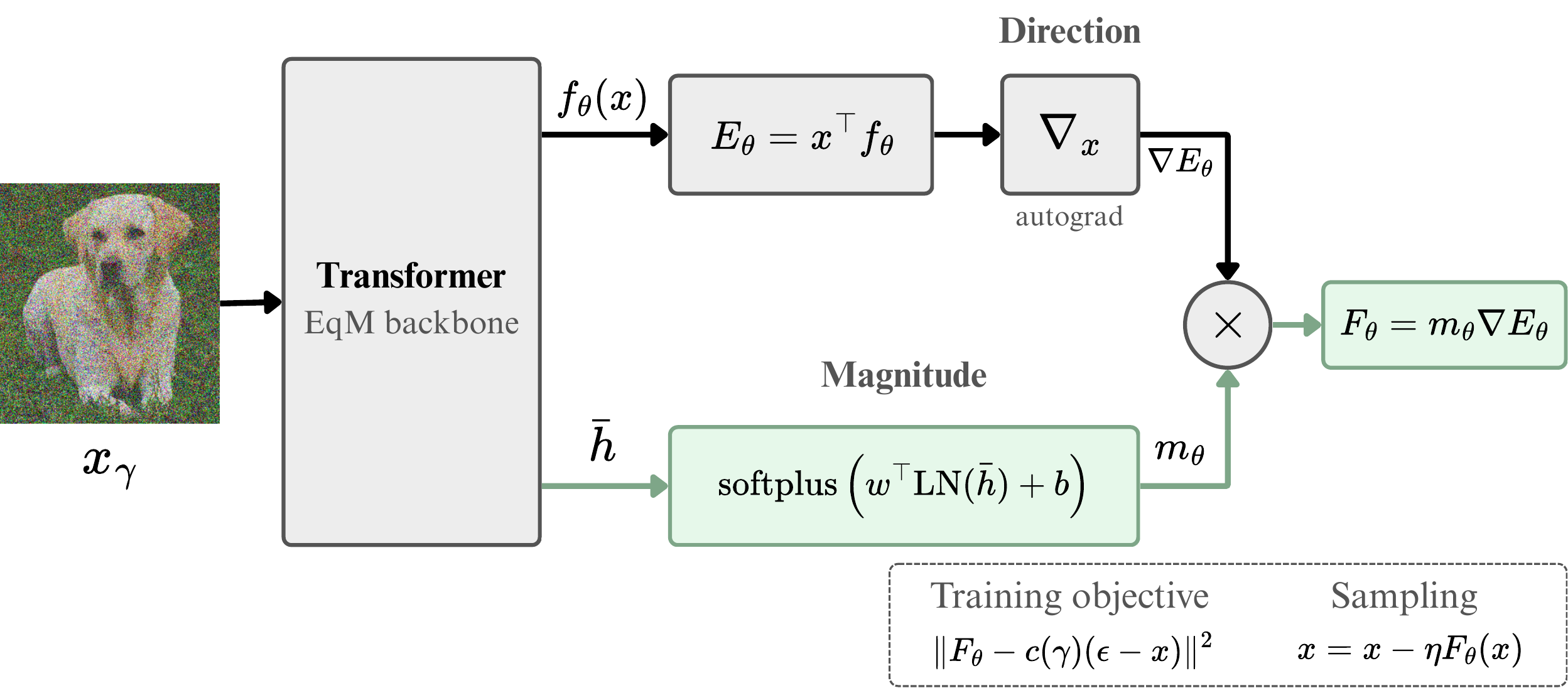}
    \caption{Overview of CEqM. The gain $m$ is learned through a separate head, multiplied by the energy's gradient $\nabla E$, allowing the field to vanish even if the energy gradient does not. The training objective and sampling stay unchanged from \eqme.}
  \label{fig:model_diagram}
\end{figure}

\begin{figure*}[t]
    \centering
    \includegraphics[width=\textwidth]{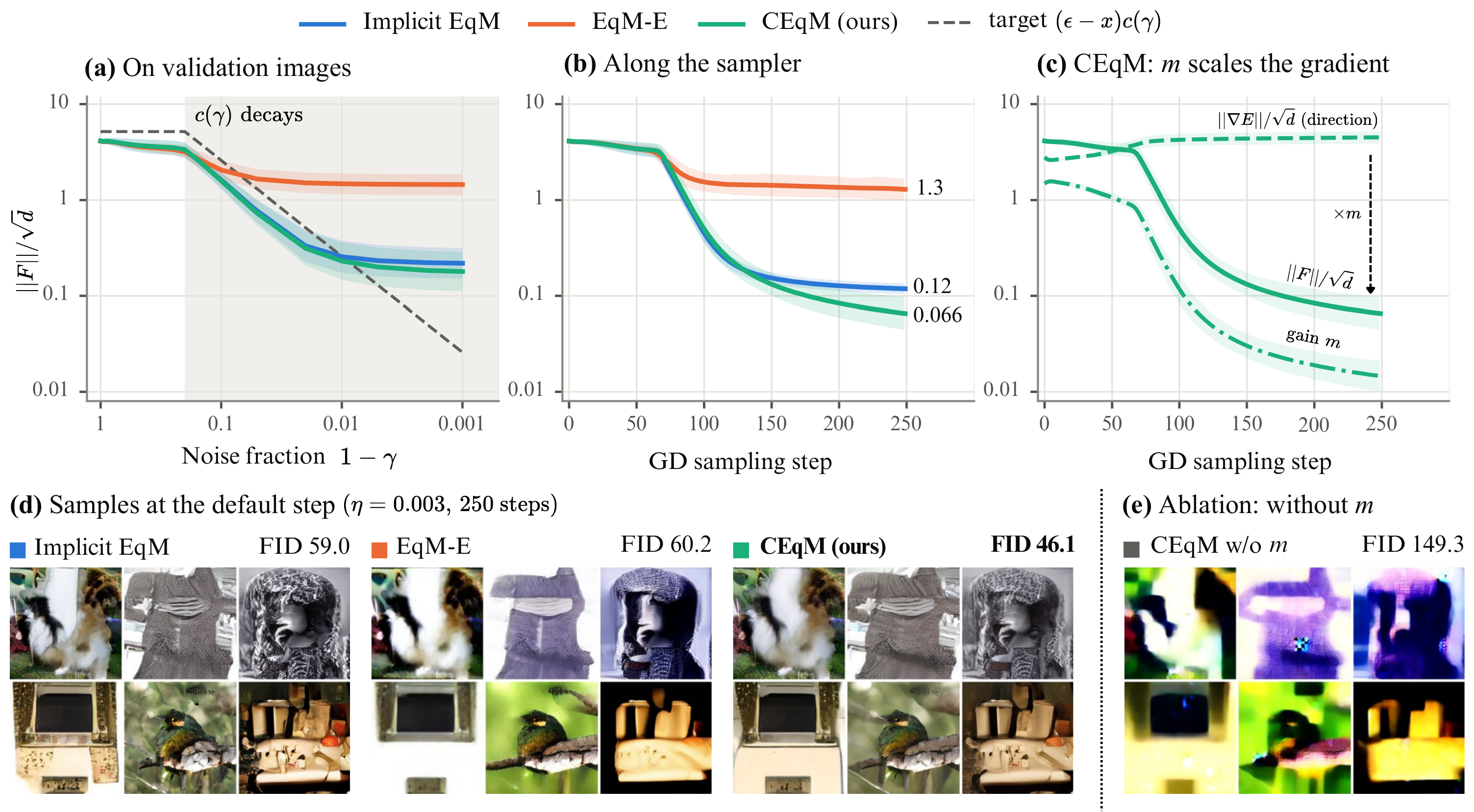}
    \caption{Normalized field norm $\|\bm{F}\|/\sqrt{d}$ of EqM-B/4 models on ImageNet. Curves show the geometric mean and bands $\pm1$ geometric standard deviation. \textbf{(a)} On 1,000 ImageNet validation images interpolated with Gaussian noise. \textbf{(b)} Along the gradient-descent trajectories of 256 samples; numbers give the final values. \textbf{(c)} CEqM's field on the same trajectories, decomposed into the energy gradient $\|\nabla E\|/\sqrt{d}$ (dashed), which plateaus and does not decay, and the gain $m$ (dash-dot), which carries the decay. \textbf{(d)} Samples and FID at the default number of sampling steps. \textbf{(e)} Removing the gain $m$ from \ours{} severely degrades samples.}
    \label{fig:field_norms}
\end{figure*}

\subsection{Formulation: Separating Direction From Magnitude} \label{sec:method_direction_from_magnitude}
\Cref{fig:model_diagram} shows the overview of \ours{}.
Using a network $\bm{f}_\theta$ with parameters $\theta$, we define the field $\bm{F}$ for input $\x$ as:
\begin{equation}
  \bm{F}(\x)=\m(\x)\nabla_{\x}\E(\x),\qquad \m(\x)>0,
  \label{eq:factorized_field}
\end{equation}
with the dot-product energy $E_\theta(\x)=\x^\top\bm{f}_\theta(\x)$, where $\nabla\E$ supplies the direction and $\m$ the magnitude. Because $\m$ can vary along the level sets of $\E$, Eq.~\eqref{eq:factorized_field} is not restricted to gradient fields, and the field can vanish where $\m\to0$ even if $\nabla\E$ does not. The gain is predicted by a small head on the average $\bar{\bm{h}}$ of the final transformer tokens (or feature map),
\begin{equation}
  \m(\x) = \mathrm{softplus}\big(\bm{w}^\top \mathrm{LN}(\bar{\bm{h}}) + b\big),
  \label{eq:gain}
\end{equation}
where LN is a layer normalization without learnable parameters, so the head has $d_h+1$ parameters for hidden width $d_h$. We initialize $\bm{w}=\bm{0}$ and $b=\log(e-1)$, so that $\m\equiv1$ and the model starts as \eqme{}. 

\Cref{tab:classes} summarizes the difference between the original EqM variants and CEqM: CEqM includes an explicit energy and nonconservative fields. The field class of Eq.~\eqref{eq:factorized_field}, and which part of the asymmetry of the optimal field it can and cannot represent, are characterized in \cref{sec:supp_class,sec:supp_structure}.

\subsection{Why the Scalar Helps}
The product rule gives
\begin{equation}
 \bm{J}_{\bm{F}}-\bm{J}_{\bm{F}}^\top
 =\nabla E(\nabla m)^\top-\nabla m(\nabla E)^\top.
 \label{eq:skew}
\end{equation}
Unlike a pure gradient, this field can have an asymmetric Jacobian. The freedom comes from changing the scalar across inputs; a constant multiplier would not help. The proof of Eq.~\eqref{eq:skew}, the rank of this asymmetry, and the non-uniqueness of the pair $(\E,\m)$ are given in Appendix~\ref{sec:supp_class}.

\subsection{Noise-Level Estimation Perspective}
When we use additive noise and analyze \ours{} from the perspective of noise-level estimation, the role of the scalar gain $\m$ becomes clear. Let $\bm{z}=\x+\sigma\eps$ with an unobserved noise scale $\sigma$, and learn the residual $\bm{z}-\x$ without observing $\sigma$. Let $q$ be the marginal density of $\bm{z}$ and $S$ its variance-weighted counterpart, obtained by weighting the density at each noise level by $\sigma^2$. Then we have
\begin{equation}
 \Ex[\bm{z}-\x\mid\bm{z}]=m^\star(\bm{z})\,\nabla E^\star(\bm{z}),
 \label{eq:exact_case}
\end{equation}
where $E^\star(\bm{z})=-\log S(\bm{z})$ and $m^\star(\bm{z})=S(\bm{z})/q(\bm{z})=\Ex[\sigma^2\mid\bm{z}]$. This factorization follows exactly from Tweedie's formula. Thus, the scalar gain $\m$ in \ours{} can be interpreted as an estimate of the noise level made from an input corrupted by unknown noise. \Cref{sec:supp_exact} gives the definitions and proof.

\subsection{Training and Sampling}\label{sec:method_train_and_sample}
We train with the EqM objective~(\ref{eq:eqm_objective}), replacing $\bm{F}$ by the factorized field described in \cref{eq:factorized_field}:
\begin{equation}
  \mathcal{L} = \Ex_{\x,\eps,\gamma}\Big[\tfrac12\,\norm{\m(\xg)\,\nabla\E(\xg) - \bm{v}(\xg)}^2\Big].
  \label{eq:loss}
\end{equation}
During sampling, we use the same gradient descent, $\x_{k+1} = \x_k - \eta\,\m(\x_k)\,\nabla\E(\x_k)$, where the gradient is given by differentiating the energy with respect to $\x_k$. The gain acts as a learned, per-sample step size. \Cref{sec:supp_descent} shows that $\E$ decreases along the resulting dynamics and gives a step-size condition for the discrete sampler, and \cref{sec:supp_geometry} identifies $\m\nabla\E$ with the Riemannian gradient of $\E$ under the conformal metric $\m^{-1}\bm{I}$.

\begin{table}[t] 
    \centering
    \small
    \setlength{\tabcolsep}{3pt}
    \caption{The three model classes. \ours{} can represent part of the rotation of the optimal field in high dimensions while preserving a scalar energy that decreases as the sampler progresses.}
    \label{tab:classes}
    \begin{threeparttable}
    \begin{tabular*}{\tabstretch\columnwidth}{@{\extracolsep{\fill}}lccc@{}}
        \toprule
    Model & Field & Rotation $\bm{A}_{\bm{F}}$ & Energy \\
        \midrule
        Implicit EqM & $\bm{f}_\theta$ & Unrestricted & \xmark \\
        \eqme & $\nabla E$ & None & \cmark \\
        \textbf{\ours{} (ours)} & $m\nabla E$ & Structured & \cmark \\
        \bottomrule
  \end{tabular*}
  \end{threeparttable}
\end{table}

\section{Experiments}
\subsection{Setup}\label{sec:setup}
We train class-conditional models on ImageNet-1K~\cite{deng2009imagenet} at $256{\times}256$ compressed in the latent space~\cite{rombach2022high} ($32{\times}32{\times}4$). All variants use the architecture and hyperparameters of EqM~\cite{wang2025eqm}: AdamW with learning rate $10^{-4}$, no weight decay, batch size 256, EMA decay 0.9999, and the truncated schedule with $a=0.8$ and $\lambda=4$. EqM-B/4 models are trained for 400k iterations ($\approx$80 epochs). For EqM-B/2, we use the official implicit EqM checkpoint~\cite{wang2025eqm}, trained for 80 epochs, and train the two explicit-energy variants from scratch with the same hyperparameters. Because of the double backward pass, a training iteration of the explicit variants takes about three times as long as one of implicit EqM in our runs. We sample with plain gradient descent ($\eta=0.003$) for 250 steps, with EMA weights and no guidance. For CIFAR-10 and CIFAR-100, all three variants use the unconditional U-Net of \cite{lipman2023flow} and its optimizer (AdamW, learning rate $10^{-4}$, $\beta = (0.9, 0.95)$, weight decay 0.01, EMA decay 0.999, batch size 256), the piecewise schedule of~\cite{wang2025eqm} with $a = 0.8$, $b = 0.4$, and $\lambda = 2$, and 200k iterations ($\approx$1,000 epochs). We sample them with Heun's method for 50 steps over $[0, 1]$. We report FID-50K~\cite{heusel2017gans}, sFID~\cite{nash2021generating}, Inception Score~\cite{salimans2016improved}, and precision and recall~\cite{kynkaanniemi2019improved}, computed with the ADM evaluation suite~\cite{dhariwal2021diffusion}. All results are from one training seed.

\subsection{Generation Quality} \label{sec:generation_quality}
\paragraph{ImageNet-1K Dataset} 
\Cref{tab:generation_imagenet} summarizes the results. With EqM-B/4, \ours{} attains the lowest FID~(46.07), against 60.18 for \eqme{} and 59.0 for implicit EqM. \ours{} also attains the highest IS, precision, and recall. With EqM-B/2, \ours{} achieves an FID of 34.69, while \eqme{} falls to 44.93.    

\paragraph{CIFAR Datasets}
\Cref{tab:generation_cifar} summarizes the unconditional U-Net results. On CIFAR-10, \ours{} lowers the FID of \eqme{} from 8.20 to 5.32 and attains the best sFID (4.82), shrinking the FID gap to implicit EqM from 4.35 to 1.47. CIFAR-100 follows the same pattern: \ours{} lowers the FID of \eqme{} from 11.43 to 7.37, shrinking the gap to implicit EqM from 6.73 to 2.67, and matches its sFID (5.18 \vs 5.17). On both datasets, \ours{} also improves IS and precision over \eqme{}, although implicit EqM remains best in FID, IS, and precision.

\begin{table}[t]
\centering
\footnotesize
\setlength{\tabcolsep}{3pt}
\caption{ImageNet-1K generation results. FID-50K, sFID, Inception Score, precision, and recall from the ADM evaluation suite. All models are trained for 400k iterations~($\approx$80 epochs) and sampled by gradient descent~($\eta = 0.003$, 250 steps) with EMA weights and no guidance. At each model size, the three variants share the architecture and training recipe. \ours{} lowers both the FID and the sFID of EqM-E at both sizes. Best per column and model size in \textbf{bold}.}
\label{tab:generation_imagenet}
\begin{threeparttable}
\begin{tabular*}{\tabstretch\columnwidth}{@{\extracolsep{\fill}}lccccc@{}}
\toprule
Model & FID $\downarrow$ & sFID $\downarrow$ & IS $\uparrow$ & Precision $\uparrow$ & Recall $\uparrow$ \\
\midrule
\multicolumn{6}{l}{\emph{EqM-B/4}} \\
Implicit EqM & 59.00 & 12.05 & 23.73 & 0.40 & 0.58 \\
\eqme{}        & 60.18 & 22.77 & 22.34 & 0.30 & 0.59 \\
\textbf{\ours{}~(ours)}       & \textbf{46.07} & \textbf{8.13} & \textbf{32.32} & \textbf{0.45} & \textbf{0.64} \\
\midrule
\multicolumn{6}{l}{\emph{EqM-B/2}} \\
Implicit EqM\tnote{$\dagger$} & \textbf{33.22} & 7.94 & \textbf{43.98} & \textbf{0.54} & \textbf{0.64} \\
\eqme{}        & 44.93 & 18.90 & 30.86 & 0.37 & 0.63 \\
\textbf{\ours{}~(ours)}        & 34.69 & \textbf{6.61} & 42.15 & 0.52 & \textbf{0.64} \\
\bottomrule
\end{tabular*}
\begin{tablenotes}[flushleft]
\scriptsize
\item[$\dagger$] The official EqM-B/2 checkpoint, evaluated in our pipeline.
\end{tablenotes}
\end{threeparttable}
\end{table}

\begin{table}[t]
    \centering
    \footnotesize
    \setlength{\tabcolsep}{3pt}
    \caption{Unconditional CIFAR generation results. All models use the U-Net of \cite{lipman2023flow}, trained for 200k iterations~($\approx$1000 epochs), and are sampled with Heun's method (50 steps). Metrics are as in \cref{tab:generation_imagenet}. \ours{} lowers both the FID and the sFID of EqM-E on both datasets. Best per column and dataset in \textbf{bold}.}
    \label{tab:generation_cifar}
    \begin{tabular*}{\tabstretch\columnwidth}{@{\extracolsep{\fill}}lccccc@{}}
        \toprule
        Model & FID $\downarrow$ & sFID $\downarrow$ & IS $\uparrow$ & Precision $\uparrow$ & Recall $\uparrow$ \\
        \midrule
        \multicolumn{6}{@{}l}{\emph{CIFAR-10, U-Net}} \\
        Implicit EqM & \textbf{3.85} & 5.16          & \textbf{9.68} & \textbf{0.652} & 0.599 \\
        \eqme{}        & 8.20          & 5.21          & 8.94          & 0.576          & \textbf{0.618} \\
        \textbf{\ours{}~(ours)}        & 5.32          & \textbf{4.82} & 9.21          & 0.631          & 0.608 \\
        \midrule
        \multicolumn{6}{@{}l}{\emph{CIFAR-100, U-Net}} \\
        Implicit EqM & \textbf{4.70} & \textbf{5.17} & \textbf{11.38} & \textbf{0.634} & 0.610 \\
        \eqme{}        & 11.43         & 5.91          & 9.84           & 0.555          & 0.609 \\
        \textbf{\ours{}~(ours)}         & 7.37          & 5.18          & 10.00          & 0.615          & \textbf{0.617} \\
        \bottomrule
    \end{tabular*}
\end{table}

\begin{table*}[t]
  \centering
  \small
  \caption{OOD detection with CIFAR-10 as in-distribution: AUROC with OOD images as positives. Top: values reported by Wang and Du~\cite{wang2025eqm} for \eqme{}-B/4, with the baselines they adopt from Yoon et al.~\cite{yoon2023energy}. Bottom: our models, trained on CIFAR-10 and scored on its test images: unconditional U-Nets (200k iterations) and class-conditional EqM-B/4 models on pixels (400k iterations, ImageNet recipe with the U-Nets' piecewise schedule), scored with the null label. Best per column in \textbf{bold}.}
\label{tab:ood_cifar}
  \begin{threeparttable}
  \begin{tabular*}{\tabstretch\textwidth}{@{\extracolsep{\fill}}ll*{8}{c}@{}}
    \toprule
    Model & Score & SVHN & Textures & Constant & CIFAR-100 & Interp & CelebA & First 3 Avg. & All 6 Avg. \\
    \midrule
    PixelCNN++ & likelihood & 0.32 & 0.33 & 0.71 & -- & -- & -- & 0.45 & -- \\
    Glow & likelihood & 0.24 & 0.27 & -- & -- & -- & -- & -- & -- \\
    IGEBM & energy & 0.63 & 0.48 & 0.39 & -- & -- & -- & 0.50 & -- \\
    \eqme{}-B/4\tnote{$\dagger$} & $\E$ & 0.55 & 0.49 & \textbf{1.00} & -- & -- & -- & 0.68 & -- \\
    \midrule
    \multicolumn{10}{l}{\textit{U-Net, 200k steps}} \\
    Implicit EqM & $\norm{\bm{F}}$ & 0.24 & 0.59 & 0.64 & 0.59 & 0.19 & 0.63 & 0.49 & 0.48 \\
    \eqme{} & $\E$ & 0.55 & 0.61 & 0.29 & 0.46 & 0.78 & 0.36 & 0.48 & 0.51 \\
    \eqme{} & $\norm{\bm{F}}$ & 0.62 & 0.68 & 0.75 & \textbf{0.62} & 0.62 & 0.60 & 0.68 & 0.65 \\
    \ours{} & $\E$ & 0.57 & 0.61 & 0.32 & 0.45 & \textbf{0.82} & 0.33 & 0.50 & 0.52 \\
    \ours{} & $\norm{\bm{F}}$ & 0.47 & 0.79 & 0.88 & 0.58 & 0.32 & 0.55 & 0.71 & 0.60 \\
    \ours{} & $\m$ & \textbf{0.76} & \textbf{0.82} & 0.98 & 0.57 & 0.36 & 0.50 & \textbf{0.85} & \textbf{0.67} \\
    \midrule
    \multicolumn{10}{l}{\textit{EqM-B/4, 400k steps}} \\
    Implicit EqM & $\norm{\bm{F}}$ & 0.39 & 0.81 & 0.99 & 0.59 & 0.27 & \textbf{0.64} & 0.73 & 0.61 \\
    \eqme{} & $\E$ & 0.57 & 0.62 & 0.31 & 0.45 & 0.80 & 0.35 & 0.50 & 0.52 \\
    \eqme{} & $\norm{\bm{F}}$ & 0.46 & 0.50 & 0.51 & 0.55 & 0.33 & 0.63 & 0.49 & 0.50 \\
    \ours{} & $\E$ & 0.55 & 0.64 & 0.32 & 0.45 & 0.81 & 0.33 & 0.50 & 0.52 \\
    \ours{} & $\norm{\bm{F}}$ & 0.23 & 0.67 & 0.62 & 0.56 & 0.43 & 0.63 & 0.51 & 0.52 \\
    \ours{} & $\m$ & 0.40 & 0.69 & 0.84 & 0.55 & 0.51 & 0.58 & 0.65 & 0.60 \\
    \bottomrule
  \end{tabular*}
    \begin{tablenotes}[flushleft]
    \scriptsize
    \item[$\dagger$] We could not reproduce the Constant score of Wang and Du~\cite{wang2025eqm}, whose construction for this dataset is not specified.
    \end{tablenotes}
    \end{threeparttable}
\end{table*}

\begin{figure*}[t]
    \centering
    \includegraphics{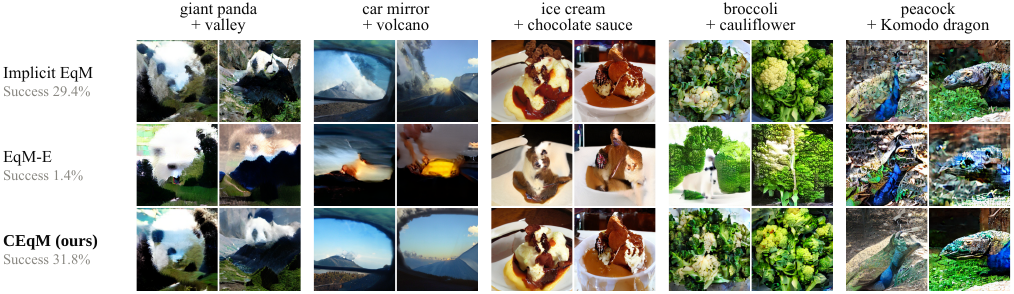}
    \caption{Compositional generation results on EqM-B/2. Curated class pairs, where the first four are from~\cite{wang2025eqm}, with the same starting noise across models. The success rate is the share of samples whose top-5 predictions from a pretrained ResNet-50 classifier contain both classes, over 50 random class pairs with 50 samples each.}
  \label{fig:composition_curated}
\end{figure*}

\subsection{How Does \ours{} Brake?} \label{sec:stop}
The field of Eq.~\eqref{eq:factorized_field} vanishes where $\m\to0$ or where $\nabla\E=0$, and \ours{} uses the former. We measure the normalized field norm $\|\bm{F}\|/\sqrt{d}$ on 1{,}000 ImageNet validation images corrupted at noise fractions $1-\gamma$ from $1$ to $10^{-3}$~(Fig.~\ref{fig:field_norms}a), and along the gradient-descent trajectories of 256 samples drawn from noise, with class labels shared across variants~(Fig.~\ref{fig:field_norms}b). As the noise fraction approaches $10^{-3}$, the fields of implicit EqM and \ours{} shrink to 0.22 and 0.18, whereas that of \eqme{} stops shrinking soon after the target starts to decay and plateaus at 1.44. None of the three reaches the norm of the training target, so even for implicit EqM, the field vanishes only approximately; \eqme{}'s field, however, stays 6--8$\times$ larger and does not behave equilibrium-like.

Along the sampler, \ours{}'s field vanishes through $\m$ rather than by reaching critical points of $\E$: $\|\nabla\E\|/\sqrt{d}$ rises from about 2.6 to 4.5 and then stays flat, while $\m$ falls from about 1.5 to 0.015~(Fig.~\ref{fig:field_norms}c). This decay starts abruptly near step 70, when the field norms of all three variants begin to fall~(Fig.~\ref{fig:field_norms}b); in terms of energy, the samples begin to slow down once they reach the median energy of real validation images~(Fig.~\ref{fig:teaser}). Sampling the trained \ours{} with $\nabla\E$ alone, from the same noise, does not stop at the data: FID rises from 46.1 to 149.3, and the samples are oversaturated like those of \eqme{} after extra steps~(Figs.~\ref{fig:teaser}b and \ref{fig:field_norms}e), suggesting that both explicit energies keep decreasing past the data. Because removing $\m$ also removes the step-size decay, this ablation does not tell whether $\m$ helps by shaping $\E$ during training or only by braking during sampling. Consistent with this braking, \ours{} is also robust to the step size: over $\eta \in [0.0015, 0.006]$, its FID varies by 1.8 at EqM-B/4 and 0.8 at EqM-B/2, as little as implicit EqM (2.3 and 0.9), against 43.6 and 51.1 for \eqme{}~(Appendix~\ref{sec:supp_eta}).

\subsection{Compositional Generation} \label{sec:composition}
We can use the explicit energy to compose conditions by adding them together~\cite{du2023reduce, liu2022compositional}. For two classes $y_1,y_2$, we descend $E_c$:
\begin{equation}
    E_c=E+w(E_{y_1}-E)+w(E_{y_2}-E),
    \label{eq:compose_conjunction}
\end{equation}
where $E$ is the energy with the null label, and weight $w$ controls guidance strength. We set $w=2.0$ for all energy composition experiments. For implicit EqM, we combine the fields in the same way as in~\cite{wang2025eqm}. We draw inspiration from~\cite{liu2022compositional} to evaluate the energy composition of each trained model. We use a ResNet-50 classifier pretrained on ImageNet and consider a sample successful if both classes appear in its top-5 predictions. Over 50 random class pairs, the success rates of \ours{}, implicit EqM, and \eqme{} are 24.2\%, 12.5\%, and 1.8\% on EqM-B/4, and 31.8\%, 29.4\%, and 1.4\% on EqM-B/2. In hand-picked pairs~(Fig.~\ref{fig:composition_curated}), \eqme{}'s samples look smeared and unrecognizable, while \ours{} is able to compose energies similarly to implicit EqM. The explicit energy of \ours{} is thus usable for composition, unlike that of EqM-E. However, the implicit EqM's field becomes comparable when the model is scaled to EqM-B/2.

\subsection{OOD Detection}
We evaluate OOD detection on CIFAR-10, using it as the ID dataset. As for OOD data, we use three prior datasets (SVHN, Textures, and Constant) and compare our results with the values reported in \cite{wang2025eqm} and the baselines they cite from \cite{yoon2023energy}. We also use three additional sets: CIFAR-100, CelebA, and Interp (the average of two CIFAR-10 test images). We score each image with different components of the trained model: field norm $\norm{\bm{F}}$, the energy $\E$, and the gain $\m$. In \Cref{tab:ood_cifar}, we report AUROC with OOD images as positives. On the three sets of prior work, the energy $E_\theta$ is near chance on average for both explicit models~(0.48 and 0.50) and inverted on Constant~(0.29--0.32). With the U-Net, the gain $m_\theta$ reaches 0.85 on these sets, above the 0.68 reported for \eqme{}, whose Constant score we could not reproduce; with EqM-B/4 it reaches 0.65, below the field norm of implicit EqM~(0.73).

\section{Conclusion} \label{sec:conclusion}
We revisited why explicit-energy variants of EqM fall short of their implicit counterparts and proposed conformal EqM (\ours{}). Our analysis traces this gap to a structural mismatch between the EqM objective's field and an energy gradient field. Encoding the magnitude and direction into a single scalar energy function degrades the representation of such a field. \ours{} decomposes the field into a positive gain and a scalar energy, which recovers the practical benefits of implicit EqM, while retaining an explicit energy that becomes useful downstream, such as composing conditions. Because a scalar gain captures only a part of the field's asymmetry, \ours{} does not uniformly close the gap to implicit models, which motivates richer factorizations. Finally, it would be interesting to further study the learned gain as it behaves like an estimate of the noise level. This suggests that it could directly serve as a noise estimator or as an adaptive step size. Validating this at a larger scale is also an important direction for future work.

\appendices
\crefalias{section}{appendix}
\crefalias{subsection}{subappendix}
\newcommand{\bz}{\bm{z}}
\newcommand{\bff}{\bm{F}}
\newcommand{\bu}{\bm{u}}
\newcommand{\bv}{\bm{v}}
\newcommand{\bw}{\bm{w}}
\newcommand{\bg}{\bm{g}}
\newcommand{\bI}{\bm{I}}
\newcommand{\bJ}[1]{\bm{J}_{#1}}
\newcommand{\bA}[1]{\bm{A}_{#1}}
\newcommand{\bS}[1]{\bm{S}_{#1}}
\newcommand{\gammahat}{\hat{\gamma}}
\newcommand{\curl}{\operatorname{curl}}
\newcommand{\grad}{\operatorname{grad}}
\newcommand{\pdata}{p_{\mathrm{data}}}
\newenvironment{restatement}[1]{\par\medskip\noindent\textbf{#1 (restated).}\itshape\ }{\par\medskip}

\section{Notation and Assumptions}
\label{sec:supp_notation}

\Cref{sec:supp_symmetry,sec:supp_optimal,sec:supp_gap,sec:supp_structure,sec:supp_class,sec:supp_exact,sec:supp_descent,sec:supp_geometry} contain the precise statements and proofs behind \cref{sec:preliminaries,sec:diagnosis,sec:method}.
\Cref{sec:supp_symmetry} characterizes gradient fields by Jacobian symmetry; \cref{sec:supp_optimal} derives the optimal EqM field and its antisymmetric part; \cref{sec:supp_gap} proves that \eqme{} has a strictly positive approximation gap; \cref{sec:supp_structure} describes the structure of the asymmetry; \cref{sec:supp_class} characterizes the fields of \ours{}; \cref{sec:supp_exact} proves the additive-noise factorization of Eq.~\eqref{eq:exact_case}; \cref{sec:supp_descent} proves energy descent along the sampler; and \cref{sec:supp_geometry} gives the Riemannian reading of the gain.

\paragraph{Notation}
$\x\in\R^D$ is a data sample with distribution $\pdata$, $\eps\sim\mathcal{N}(\bm{0},\bI)$, and $\gamma\sim\pi$ is independent of both, with $\pi$ the uniform distribution on $[0,1]$ unless stated otherwise; $\xg=\gamma\x+(1-\gamma)\eps$ as in \cref{sec:preliminaries}.
$\bz$ denotes a generic network input.
$p_\gamma$ is the density of $\xg$ at fixed $\gamma$, $q(\bz)=\int_0^1p_\gamma(\bz)\,\pi(d\gamma)$ is the marginal density of the training inputs, and $p(\gamma\mid\bz)=p_\gamma(\bz)\pi(\gamma)/q(\bz)$ is the posterior over the interpolation factor.
The schedule $c:[0,1]\to[0,\infty)$ is piecewise $C^1$ with $c(1)=0$; the truncated schedule $c(\gamma)=\lambda\min\{1,(1-\gamma)/(1-a)\}$ of \cref{sec:setup} is the main example.
Expectations are written $\Ex$; $\E$ and $\m$ denote the learned energy and gain.
Jacobians are $(\bJ{\bff})_{ij}=\partial_jF_i$, with antisymmetric part $\bA{\bff}=\tfrac12(\bJ{\bff}-\bJ{\bff}^{\top})$ as in Eq.~\eqref{eq:antisym} and symmetric part $\bS{\bff}=\tfrac12(\bJ{\bff}+\bJ{\bff}^{\top})$.
Sign convention: the field is regressed onto $c(\gamma)(\eps-\x)$, which points from the data towards the noise, and the sampler is $\x_{k+1}=\x_k-\eta\bff(\x_k)$, so the energy decreases towards the data.
A class condition, when used, is omitted from the notation.

\paragraph{Assumptions}
\begin{itemize}
\item[(A1)] $\pdata$ has finite second moment.
Then for every $\gamma\in[0,1)$ the density $p_\gamma$ exists, is $C^\infty$, and is strictly positive, being a Gaussian smoothing of the law of $\gamma\x$ at scale $1-\gamma$.
\item[(A2)] Derivatives of integrals over $\gamma$ may be taken under the integral sign (dominated convergence). This holds, for instance, if $\pi$ is supported in $[0,1-\delta]$ for some $\delta>0$, or if $\pdata$ is a nondegenerate Gaussian mixture, as in \cref{ex:supp_2d}. A bounded density alone does not suffice: at the boundary of the support of $\pdata$, $\norm{\nabla p_\gamma}$ grows like $(1-\gamma)^{-1}$, so $\nabla q$ diverges there, and $c(1)=0$ does not help for the marginal $q$.
\item[(A3)] Fields are $C^1$ where their Jacobians are used, and energies are $C^2$ where their Hessians are used.
\end{itemize}

\section{Gradient Fields and Jacobian Symmetry}
\label{sec:supp_symmetry}

\begin{proposition}[Conservative fields have symmetric Jacobians, and conversely]
\label{prop:supp_symmetry}
Let $\bff:\R^D\to\R^D$ be $C^1$.
Then $\bff=\nabla E$ for some $E\in C^2(\R^D)$ if and only if $\bJ{\bff}(\bz)$ is symmetric for every $\bz$, \ie, $\bA{\bff}\equiv0$.
\end{proposition}
\begin{proof}
If $\bff=\nabla E$ with $E\in C^2$, then $\bJ{\bff}=\nabla^2E$ is symmetric by Schwarz's theorem.
Conversely, symmetry means $\partial_iF_j=\partial_jF_i$, \ie, the one-form $\sum_iF_i\,dz_i$ is closed.
Since $\R^D$ is star-shaped, the Poincar\'e lemma gives a $C^2$ function $E$ with $dE=\sum_iF_i\,dz_i$, explicitly $E(\bz)=\int_0^1\bff(t\bz)\cdot\bz\,dt$.
\end{proof}

The converse uses that the domain is all of $\R^D$ (or any simply connected open set); on a domain with holes, a closed field need not be exact. For $D=2$, $\bA{\bff}=\tfrac12(\partial_1F_2-\partial_2F_1)\begin{psmallmatrix}0&-1\\1&0\end{psmallmatrix}$, so $\bA{\bff}$ is the scalar curl shown in \cref{fig:toy_example}; for $D=3$ its independent entries are the components of $\nabla\times\bff$.

Two clarifications of the informal language of \cref{sec:diagnosis} are in order.
First, $\bA{\bff}$ is a \emph{pointwise} test of whether $\bff$ is a gradient; the $L^2(q)$ distance of $\bff$ to the nearest gradient field (\cref{sec:supp_gap}) is a global quantity, and neither determines the other.
Second, $\bA{\bff}\neq0$ rules out that $\bff$ is the gradient of \emph{some} energy; it does not rule out that $\bff$ descends an energy.
The fields of \ours{} have $\bA{\bff}\neq0$ in general (\cref{sec:supp_class}) and still decrease $\E$ along their flow (\cref{sec:supp_descent}).

\section{The Optimal Equilibrium Field}
\label{sec:supp_optimal}

\paragraph{Conditional expectation}
For any measurable $\bff$ with $\Ex\norm{\bff(\xg)}^2<\infty$, the population loss of Eq.~\eqref{eq:eqm_objective} decomposes as
\begin{equation}
\begin{aligned}
  \mathcal{L}_{\mathrm{EqM}}(\bff)
  &=\tfrac12\Ex\norm{\bff(\xg)-\bff^\star(\xg)}^2 \\
  &+\tfrac12\Ex\norm{\bff^\star(\xg)-\bm{v}(\xg)}^2,
  \label{eq:supp_decomp}
\end{aligned}
\end{equation}
with $\bm{v}(\xg)=c(\gamma)(\eps-\x)$ and
\begin{equation}
\begin{aligned}
  \bff^\star(\bz)
  &=\Ex\big[c(\gamma)(\eps-\x)\,\big|\,\xg=\bz\big]\\
  &=\int_0^1p(\gamma\mid\bz)\,c(\gamma)\,\Ex\big[\eps-\x\,\big|\,\bz,\gamma\big]\,d\gamma ,
\end{aligned}
  \label{eq:supp_fstar}
\end{equation}
which is Eq.~\eqref{eq:fstar}.
Hence $\bff^\star$ is the unique minimizer up to $q$-null sets, and the excess loss of any field is half its squared $L^2(q)$ distance to $\bff^\star$.
The second equality in Eq.~\eqref{eq:supp_fstar} is the tower property with respect to $\gamma$.

\begin{lemma}[Per-level targets are gradient fields]
\label{lem:supp_tweedie}
For every fixed $\gamma\in(0,1)$,
\begin{equation}
\begin{aligned}
  \Ex[\eps-\x\mid\bz,\gamma]&=\nabla_{\bz}\Psi_\gamma(\bz),\\
  \Psi_\gamma(\bz)&\!=\!-\frac1\gamma\Big(\tfrac12\norm{\bz}^2\!+\!(1\!-\!\gamma)\log p_\gamma(\bz)\Big).
\end{aligned}
  \label{eq:supp_psi}
\end{equation}
\end{lemma}
\begin{proof}
Given $\gamma$ and $\x$, $\bz\sim\mathcal{N}(\gamma\x,(1-\gamma)^2\bI)$.
Tweedie's formula~\cite{efron2011tweedie} for Gaussian smoothing at scale $1-\gamma$ gives $\Ex[\gamma\x\mid\bz,\gamma]=\bz+(1-\gamma)^2\nabla\log p_\gamma(\bz)$.
Hence $\Ex[\eps\mid\bz,\gamma]=(\bz-\Ex[\gamma\x\mid\bz,\gamma])/(1-\gamma)=-(1-\gamma)\nabla\log p_\gamma(\bz)$ and $\Ex[\x\mid\bz,\gamma]=\gamma^{-1}\big(\bz+(1-\gamma)^2\nabla\log p_\gamma(\bz)\big)$, so that
\begin{equation*}
\begin{aligned}
  \Ex[\eps-\x\mid\bz,\gamma]
  &=-\frac{\bz}{\gamma}-\Big(1-\gamma+\frac{(1-\gamma)^2}{\gamma}\Big)\nabla\log p_\gamma(\bz)\\
  &=-\frac1\gamma\big(\bz+(1-\gamma)\nabla\log p_\gamma(\bz)\big)\\
  &=\nabla\Psi_\gamma(\bz).
\end{aligned}
\end{equation*}
\end{proof}

At $\gamma=0$ the input $\bz=\eps$ is independent of $\x$, so $\Ex[\eps-\x\mid\bz,0]=\bz-\Ex[\x]$ is also a gradient.
As $\gamma\to1$ the factor $1/\gamma$ is harmless, but $\log p_\gamma$ may lose smoothness if $\pdata$ is singular; this is covered by (A2).

\begin{proposition}[The optimal field is a mixture of gradients and need not be conservative]
\label{prop:supp_mixture}
Under (A1)--(A2), the optimal field is the input-weighted mixture
\begin{equation}
  \bff^\star(\bz)=\int_0^1p(\gamma\mid\bz)\,c(\gamma)\,\nabla\Psi_\gamma(\bz)\,d\gamma ,
  \label{eq:supp_mixture}
\end{equation}
and the antisymmetric part of its Jacobian is
\begin{equation}
\begin{aligned}
  \bA{\bff^\star}(\bz)=\frac12\int_0^1c(\gamma)\Big(
  &\nabla\Psi_\gamma(\bz)\,\nabla_{\bz}p(\gamma\mid\bz)^{\!\top}\\
  &\!-\!\nabla_{\bz}p(\gamma\!\mid\!\bz)\nabla\Psi_\gamma(\bz)^{\!\top}\!\Big)\,d\gamma .
\end{aligned}
  \label{eq:supp_Astar}
\end{equation}
Consequently:
(i) $\bff^\star$ is conservative if and only if Eq.~\eqref{eq:supp_Astar} vanishes identically;
(ii) for $c(\gamma)>0$, the integrand at level $\gamma$ vanishes if and only if $\nabla_{\bz}p(\gamma\mid\bz)$ is parallel to $\nabla\Psi_\gamma(\bz)$ (or one of them is zero), so it suffices for conservativity that every weight changes only along its own per-level direction;
(iii) the integrand can be nonzero at every level while the integral cancels, so (ii) is sufficient but not necessary.
\end{proposition}
\begin{proof}
Eq.~\eqref{eq:supp_mixture} is Eq.~\eqref{eq:supp_fstar} with \cref{lem:supp_tweedie}.
Differentiating under the integral (A2) with the product rule,
\begin{equation*}
\begin{aligned}
  \bJ{\bff^\star}
  ={}&\int_0^1p(\gamma\mid\bz)\,c(\gamma)\,\nabla^2\Psi_\gamma(\bz)\,d\gamma\\
  &+\int_0^1c(\gamma)\,\nabla\Psi_\gamma(\bz)\,\nabla_{\bz}p(\gamma\mid\bz)^{\top}d\gamma .
\end{aligned}
\end{equation*}
The first integral is a non-negative mixture of Hessians and is symmetric.
The antisymmetric part of a rank-one matrix $\bm{a}\bm{b}^{\top}$ is $\tfrac12(\bm{a}\bm{b}^{\top}-\bm{b}\bm{a}^{\top})$, which gives Eq.~\eqref{eq:supp_Astar}.
(i) is \cref{prop:supp_symmetry}.
(ii) $\bm{a}\bm{b}^{\top}-\bm{b}\bm{a}^{\top}=0$ if and only if $\bm{a}$ and $\bm{b}$ are linearly dependent.
(iii) Isotropic Gaussian data (\cref{ex:supp_2d}) make every integrand vanish.
For cancellation, let $\pdata=\delta_{\bm a}$ with $\bm a=(1,0)$, $\pi=\tfrac12(\delta_{1/4}+\delta_{1/2})$, and $c(\gamma)=1-\gamma$.
Then $c(\gamma)(\eps-\x)=\bz-\bm a$ identically, so $\bff^\star(\bz)=\bz-\bm a$ and $\bA{\bff^\star}\equiv0$, whereas $\nabla\Psi_\gamma=(\bz-\bm a)/(1-\gamma)$ and $\nabla_{\bz}p(\tfrac14\mid\bz)=-\nabla_{\bz}p(\tfrac12\mid\bz)$, which at $\bz=(0,1)$ is not parallel to $\bz-\bm a$; the two terms of the sum that replaces the integral in Eq.~\eqref{eq:supp_Astar} are therefore nonzero and cancel.
\end{proof}

\paragraph{Remarks}
(a) The dependence of $p(\gamma\mid\bz)$ on $\bz$ is unavoidable: $p(\gamma\mid\bz)\propto p_\gamma(\bz)$, and the densities at different levels have different shapes.
(b) The asymmetry is weighted by $c(\gamma)$ and therefore depends on the schedule; in \cref{ex:supp_2d} the truncated schedule produces a larger asymmetry than $c(\gamma)=1-\gamma$.
(c) Nothing in the argument involves the architecture; \cref{prop:supp_mixture} is a statement about the regression target.
(d) Time-conditional diffusion and flow models receive $\gamma$ as an input, so their optimal field at each level is the gradient of \cref{lem:supp_tweedie}~\cite{salimans2021energy}; the issue is specific to noise-unconditional fields.
(e) Eq.~\eqref{eq:supp_fstar} also makes precise in what sense real samples are stationary points (\cref{sec:preliminaries}).
Because $c(1)=0$, the \emph{target} vanishes at $\gamma=1$; the optimal field at a data point $\bz$, however, is the posterior average over all levels and vanishes only if the contributions of the levels $\gamma<1$ cancel.
This holds in the idealized case where $p(\gamma\mid\bz)$ is concentrated at $\gamma=1$ (\cref{sec:supp_structure}), but not in general: for $\x\sim\mathcal{N}(\bm 0,\bI)$, $\pi$ uniform, and $c(\gamma)=1-\gamma$, pairing the levels $\gamma$ and $1-\gamma$ in Eq.~\eqref{eq:supp_mixture} gives $\bff^\star(\bz)=\kappa(\bz)\bz$ with $\kappa(\bz)=\tfrac12\Ex\big[(1-2\gamma)^2/(\gamma^2+(1-\gamma)^2)\,\big|\,\bz\big]>0$, which is nonzero at almost every data point. The theory, therefore, does not guarantee that the optimal field is small for real data; how small the \emph{learned} fields become near real data is the empirical observation in \cref{fig:field_norms}.

\begin{example}[Two-dimensional Gaussian data]
\label{ex:supp_2d}
Let $\x\sim\mathcal{N}(\bm{0},\bm{\Sigma})$ with $\bm{\Sigma}=\operatorname{diag}(0.1,3)$ and $\gamma\sim\mathcal{U}[0,1]$.
Then $\xg\sim\mathcal{N}(\bm{0},\bm{\Sigma}_\gamma)$ with $\bm{\Sigma}_\gamma=\gamma^2\bm{\Sigma}+(1-\gamma)^2\bI$, $\nabla\log p_\gamma(\bz)=-\bm{\Sigma}_\gamma^{-1}\bz$, and by Eq.~\eqref{eq:supp_psi}
\begin{equation}
  \nabla\Psi_\gamma(\bz)=-\gamma^{-1}\big(\bI-(1-\gamma)\bm{\Sigma}_\gamma^{-1}\big)\bz .
  \label{eq:supp_gauss_psi}
\end{equation}
Eq.~\eqref{eq:supp_mixture} is then a ratio of one-dimensional integrals in $\gamma$, which we evaluate by Gauss--Legendre quadrature on $[0,a]$ and $[a,1]$ separately (the truncated schedule has a kink at $\gamma=a$); the Jacobian obtained by differentiating the integrand analytically agrees with central differences to $10^{-9}$.
At $\bz=(0.5,1)$:
\begin{center}
\small
\setlength{\tabcolsep}{4pt}
\begin{tabular}{@{}lcc@{}}
\toprule
schedule & $\curl\bff^\star$ & $\norm{\bA{\bff^\star}}_F/\norm{\bS{\bff^\star}}_F$ \\
\midrule
$c(\gamma)=1-\gamma$ & $0.290$ & $0.209$ \\
$c(\gamma)=4\min\{1,5(1-\gamma)\}$ & $2.537$ & $0.277$ \\
\bottomrule
\end{tabular}
\end{center}
Here $\curl\bff^\star=\partial_1F^\star_2-\partial_2F^\star_1$.
The anisotropy of $\bm{\Sigma}$ is essential: for $\bm{\Sigma}=\sigma^2\bI$ every $\nabla\Psi_\gamma$ is radial and $p(\gamma\mid\bz)$ depends on $\bz$ only through $\norm{\bz}$, so every integrand in Eq.~\eqref{eq:supp_Astar} vanishes and the field is conservative.
With anisotropy, the per-level directions rotate with $\gamma$ while the weights vary with $\bz$ in a direction that is not aligned with them.
The mismatch of \cref{sec:diagnosis} thus depends on the data, and its size must be measured.
\end{example}

\begin{example}[A nonconservative optimal field that \ours{} represents exactly]
\label{ex:supp_point}
Take the data and prior of \cref{prop:supp_mixture}(iii), $\pdata=\delta_{\bm a}$ with $\bm a=(1,0)$ and $\pi=\tfrac12(\delta_{1/4}+\delta_{1/2})$, but with the truncated schedule $c(\gamma)=4\min\{1,5(1-\gamma)\}$, so that $c=4$ at both levels, and let $w(\bz)=p(\tfrac12\mid\bz)$.
Since $\nabla\Psi_\gamma(\bz)=(\bz-\bm a)/(1-\gamma)$, Eq.~\eqref{eq:supp_mixture} gives
\begin{equation}
\begin{aligned}
  \bff^\star(\bz)&=\kappa(\bz)\,\nabla\tfrac12\norm{\bz-\bm a}^2,\\
  \kappa(\bz)&=\tfrac{16}{3}+\tfrac83\,w(\bz)>0,
  \label{eq:supp_point}
\end{aligned}
\end{equation}
which is exactly of the form $\m\nabla\E$ of Eq.~\eqref{eq:factorized_field}, with $q$ finite and smooth everywhere.
It is not conservative: $\nabla w=w(1-w)\nabla\log(p_{1/2}/p_{1/4})$ gives, at $\bz=(0,1)$, $\curl\bff^\star=-\tfrac{16}{9}w(1-w)\approx-0.388$.
By \cref{prop:supp_gap}, \eqme{} has a strictly positive gap on this target, while \ours{} represents it exactly; with $c(\gamma)=1-\gamma$ instead, $\kappa\equiv1$ and the field is conservative.
\end{example}

\section{The Irreducible Error of Explicit Energy Models}
\label{sec:supp_gap}

Let $\mathcal{H}=L^2(q;\R^D)$ with $\ip{\bff}{\bg}_q=\Ex_q[\bff\cdot\bg]$, and let $\mathcal{G}=\{\nabla E:E\in C^2,\ \nabla E\in\mathcal{H}\}$ be the gradient fields.
By Eq.~\eqref{eq:supp_decomp}, the best achievable excess loss of an explicit energy model is
\begin{equation}
\begin{aligned}
  \mathcal{L}_{\mathrm{cons}}
  =\inf_{E}\tfrac12\Ex_q\norm{\nabla E-\bff^\star}^2
  &=\tfrac12\operatorname{dist}_{\mathcal{H}}\big(\bff^\star,\overline{\mathcal{G}}\big)^2\\
  &=\tfrac12\Ex_q\norm{\bm{r}}^2,
  \label{eq:supp_irreducible}
\end{aligned}
\end{equation}
where $\bm{r}=\bff^\star-P_{\overline{\mathcal{G}}}\bff^\star$ and $P_{\overline{\mathcal{G}}}$ is the orthogonal projection onto the closure of $\mathcal{G}$.
The residual satisfies $\Ex_q[\bm{r}\cdot\nabla\phi]=0$ for all test functions $\phi$, \ie, $\nabla\cdot(q\bm{r})=0$ weakly: this is the $q$-weighted Helmholtz--Hodge decomposition of $\bff^\star$, and $P_{\overline{\mathcal{G}}}\bff^\star$ is the ``non-rotating part'' that \eqme{} can at best fit (\cref{sec:diagnosis}).
Two cautions: the projection need not be the gradient of a $C^2$ energy (it lies in the closure), and the decomposition is global, so the gradient part is not obtained by removing $\bA{\bff^\star}$ pointwise.

\begin{proposition}[Positive gap]
\label{prop:supp_gap}
Assume (A1)--(A3), that $q$ is continuous and strictly positive on an open set $U$, and that $\bA{\bff^\star}(\bz_0)\neq0$ for some $\bz_0\in U$.
Then $\mathcal{L}_{\mathrm{cons}}>0$.
\end{proposition}
\begin{proof}
Suppose $\mathcal{L}_{\mathrm{cons}}=0$, so $\bff^\star\in\overline{\mathcal{G}}$: there are $E_n\in C^2$ with $\nabla E_n\to\bff^\star$ in $L^2(q)$, hence in $L^2(B)$ for a ball $B\subset U$ around $\bz_0$ on which $q\ge q_{\min}>0$.
By continuity of $\bA{\bff^\star}$ there are a ball $B'\subset B$ and indices $i\neq j$ with $\partial_iF^\star_j-\partial_jF^\star_i\ge\alpha>0$ on $B'$ (after possibly swapping $i$ and $j$).
Take $\phi\in C_c^\infty(B')$ with $\phi\ge0$ and $\phi\not\equiv0$, and consider the linear functional
\begin{equation}
  \Lambda(\bff)=\int_{B'}\big(F_j\,\partial_i\phi-F_i\,\partial_j\phi\big)\,d\bz ,
  \label{eq:supp_Lambda}
\end{equation}
which is continuous on $L^2(B')$.
For $\bff=\nabla E_n$, integration by parts gives $\Lambda(\nabla E_n)=-\int_{B'}(\partial_i\partial_jE_n-\partial_j\partial_iE_n)\phi=0$, so $\Lambda(\bff^\star)=\lim_n\Lambda(\nabla E_n)=0$.
But for the $C^1$ field $\bff^\star$ the same integration by parts gives $\Lambda(\bff^\star)=-\int_{B'}(\partial_iF^\star_j-\partial_jF^\star_i)\phi\le-\alpha\int_{B'}\phi<0$, a contradiction.
\end{proof}

Thus, wherever the target has nonzero asymmetry on the support of the training inputs, \eqme{} has a strictly positive approximation gap that is independent of capacity, optimizer, and training length.
The size of the gap, and its effect on sample quality, are not determined by this argument and must be measured (\cref{sec:generation_quality,sec:stop}).

\section{Structure of the Asymmetry}
\label{sec:supp_structure}

This section justifies the entry ``structured'' for \ours{} in \cref{tab:classes}: it identifies which part of $\bA{\bff^\star}$ a positive scalar in front of a gradient can produce, and which part it cannot.

\subsection{An Exact Two-Term Decomposition}
\label{sec:supp_two_terms}

Writing $\nabla\log p_\gamma=\nabla p_\gamma/p_\gamma$ in \cref{lem:supp_tweedie} and $p(\gamma\mid\bz)=p_\gamma(\bz)\pi(\gamma)/q(\bz)$ in Eq.~\eqref{eq:supp_mixture}, the densities cancel:
\begin{equation}
\begin{aligned}
  \bff^\star(\bz)
  &=-\frac{1}{q(\bz)}\int_0^1\frac{c(\gamma)}{\gamma}\Big(p_\gamma(\bz)\,\bz+(1-\gamma)\nabla p_\gamma(\bz)\Big)\pi(d\gamma)\\
  &=-\frac{S_1(\bz)\,\bz+\nabla S_2(\bz)}{q(\bz)},
\end{aligned}
  \label{eq:supp_two_terms}
\end{equation}
with the weighted marginals
\begin{equation}
\begin{aligned}
  S_1(\bz)&=\int_0^1\frac{c(\gamma)}{\gamma}\,p_\gamma(\bz)\,\pi(d\gamma),\\
  S_2(\bz)&=\int_0^1\frac{c(\gamma)(1-\gamma)}{\gamma}\,p_\gamma(\bz)\,\pi(d\gamma).
\end{aligned}
  \label{eq:supp_S12}
\end{equation}
Here we assume $\int_0^1c(\gamma)\gamma^{-1}\,\pi(d\gamma)<\infty$, \eg, $\pi$ supported in $[\delta,1]$. For the uniform prior with $c(0)>0$, the two terms of Eq.~\eqref{eq:supp_two_terms} diverge logarithmically as $\gamma\to0$ while their sum stays finite, because $p_\gamma\bz+(1-\gamma)\nabla p_\gamma\to\varphi\bz+\nabla\varphi=0$ for the standard normal density $\varphi$; the field itself stays finite as $\delta\to0$, but the two terms do not. Since $\bz=\nabla\tfrac12\norm{\bz}^2$, Eq.~\eqref{eq:supp_two_terms} says that the optimal field is a sum of \emph{two} scalar-times-gradient fields:
\begin{equation}
  \bff^\star=-\frac{S_2}{q}\Big(\rho\,\nabla\tfrac12\norm{\bz}^2+\nabla\log S_2\Big),
  \qquad
  \rho=\frac{S_1}{S_2}.
  \label{eq:supp_rho}
\end{equation}
The ratio $\rho(\bz)=\Ex[c(\gamma)/\gamma\mid\bz]/\Ex[c(\gamma)(1-\gamma)/\gamma\mid\bz]$ is a quotient of two posterior expectations of functions of the noise level.
If $\rho$ were constant, $\bff^\star$ would be exactly of the form $\m\nabla\E$ of Eq.~\eqref{eq:factorized_field}, with $m=S_2/q>0$ and $E=-\rho\tfrac12\norm{\bz}^2-\log S_2$.
A constant $\rho$ is thus sufficient for exact representability by \ours{}; it is not necessary (see below).
Writing $\tilde m=S_2/q$ and $\bg=\rho\bz+\nabla\log S_2$, so that $\bff^\star=-\tilde m\bg$, the product rule gives
\begin{equation}
  \bA{\bff^\star}
  =-\tfrac12\big(\bg\nabla\tilde m^{\top}-\nabla\tilde m\,\bg^{\top}\big)
  +\tfrac{\tilde m}{2}\big(\nabla\rho\,\bz^{\top}-\bz\,\nabla\rho^{\top}\big),
  \label{eq:supp_A_rho}
\end{equation}
where the first term has the form of Eq.~\eqref{eq:skew} for the pair $(\tilde m,\bg)$ and the second is the asymmetry added by the variation of $\rho$. This split is not canonical, and a nonzero second term does not exclude representability: in the example of \cref{prop:supp_mixture}(iii), $\bff^\star=\bz-\bm a$ although $\nabla\rho\not\parallel\bz$, so the two terms cancel, and in \cref{ex:supp_point} $\rho$ varies while $\bff^\star$ is exactly of the form $\m\nabla\E$. The exact criterion is that of \cref{prop:supp_class}: since the first term satisfies $\bu^{\top}(\cdot)\bv=0$ for all $\bu,\bv\perp\bff^\star$, the obstruction is the component of $\nabla\rho\,\bz^{\top}-\bz\,\nabla\rho^{\top}$ in the directions orthogonal to $\bff^\star$.
When the posterior concentrates at a level $\gammahat(\bz)$, $\rho(\bz)\approx1/(1-\gammahat(\bz))$, so the second term is controlled by $\nabla\gammahat$; it plays the role of the level term of \cref{sec:supp_concentrated}. The computation also shows why the additive-noise setting of \cref{sec:supp_exact} is exactly representable: there the shrinkage factor $\gamma$ in front of $\x$ is absent, the term $S_1\bz$ does not arise, and the field is a single $-\nabla S/q$.

\subsection{The Concentrated-Posterior Approximation}
\label{sec:supp_concentrated}

In $D\gg1$ dimensions, if $\Ex\norm{\x}^2=\Theta(D)$ and $\operatorname{Var}\norm{\x}^2=O(D)$, then, since $\operatorname{Var}\norm{\eps}^2=2D$ and $\operatorname{Var}(\x\cdot\eps)=\Ex\norm{\x}^2$, the squared norm $\norm{\xg}^2=\gamma^2\norm{\x}^2+(1-\gamma)^2\norm{\eps}^2+2\gamma(1-\gamma)\x\cdot\eps$ concentrates around $\gamma^2\Ex\norm{\x}^2+(1-\gamma)^2D$ with relative fluctuations $O(D^{-1/2})$; the same holds for the energy of $\xg$ in a frequency band with $\Theta(D)$ coefficients under the analogous variance condition. Finite second moment alone does not imply this (a scale mixture of Gaussians is a counterexample). For data whose statistics differ from those of white noise, as natural images do, the posterior $p(\gamma\mid\bz)$ is then sharply peaked at a function $\gammahat(\bz)$ of the input; this identifiability is what noise-unconditional denoisers rely on~\cite{sun2025noise,mohan2020biasfree}. It is a property of the data and not a theorem: if $\x\sim\mathcal{N}(\bm{0},\bI)$, then $\xg\sim\mathcal{N}(\bm{0},(\gamma^2+(1-\gamma)^2)\bI)$, the likelihoods of $\gamma$ and $1-\gamma$ coincide in every dimension, and the level is not identifiable.

\begin{proposition}[Two sources of asymmetry]
\label{prop:supp_two_sources}
Suppose $p(\gamma\mid\bz)=\delta(\gamma-\gammahat(\bz))$ with $\gammahat$ differentiable (an idealization: for finite $D$, $p_\gamma(\bz)>0$ for all $\gamma<1$, and a peaked posterior approximates the value of $\bff^\star$, not automatically its Jacobian), and write $\bm{G}(\bz,\gamma)=\nabla_{\bz}\Psi_\gamma(\bz)$.
Then
\begin{equation}
  \bff^\star(\bz)=s(\bz)\,\bv(\bz),\quad
  s=c(\gammahat(\bz)),\quad
  \bv=\bm{G}\big(\bz,\gammahat(\bz)\big),
  \label{eq:supp_sv}
\end{equation}
and, with $\bw=\partial_\gamma\bm{G}(\bz,\gammahat(\bz))$,
\begin{equation}
\begin{aligned}
  \bA{\bff^\star}
  ={}&\underbrace{\tfrac12\big(\bv\,\nabla s^{\!\top}-\nabla s\,\bv^{\!\top}\big)}_{\text{scale term}}\\
  &+\underbrace{\tfrac{s}{2}\big(\bw\,\nabla\gammahat^{\!\top}-\nabla\gammahat\,\bw^{\!\top}\big)}_{\text{level term}} .
\end{aligned}
  \label{eq:supp_A_decomp}
\end{equation}
The scale term (a) is the complete antisymmetric part of $s\nabla\Psi$ for any fixed potential $\Psi$ with $\nabla\Psi=\bv$; (b) has rank at most two; (c) vanishes wherever $\nabla s=0$, in particular wherever $\gammahat<a$ for the truncated schedule; and (d) satisfies $\bu^{\!\top}\bA{\mathrm{scale}}\bv'=0$ for all $\bu,\bv'\perp\bff^\star(\bz)$.
The level term contains no derivative of $c$ but is multiplied by $s=c(\gammahat)$.
\end{proposition}
\begin{proof}
Substituting the point-mass posterior into Eq.~\eqref{eq:supp_mixture} gives Eq.~\eqref{eq:supp_sv}. Then $\bJ{\bff^\star}=\bv\nabla s^{\top}+s\bJ{\bv}$ and, by the chain rule, $\bJ{\bv}=\partial_{\bz}\bm{G}+\partial_\gamma\bm{G}\,\nabla\gammahat^{\top}=\nabla^2\Psi_\gamma|_{\gammahat}+\bw\nabla\gammahat^{\top}$. The Hessian is symmetric, and the antisymmetric parts of the two rank-one terms give Eq.~\eqref{eq:supp_A_decomp}.
(a) For $\bff=s\nabla\Psi$, $\bJ{\bff}=\nabla\Psi\nabla s^{\top}+s\nabla^2\Psi$.
(b) A difference of two rank-one matrices.
(c) $\nabla s=c'(\gammahat)\nabla\gammahat$ and $c'=0$ on $[0,a)$.
(d) $\bu^{\top}(\bv\nabla s^{\top}-\nabla s\bv^{\top})\bv'=(\bu\cdot\bv)(\nabla s\cdot\bv')-(\bu\cdot\nabla s)(\bv\cdot\bv')$, which vanishes when $\bu,\bv'\perp\bv\parallel\bff^\star$.
\end{proof}

\paragraph{The direction $\bv$ is not itself a gradient}
It is tempting to write $\bv=\nabla\Psi$ with $\Psi(\bz)=\Psi_{\gammahat(\bz)}(\bz)$, but the chain rule gives
\begin{equation}
  \nabla_{\bz}\big[\Psi_{\gammahat(\bz)}(\bz)\big]
  =\bv(\bz)+\partial_\gamma\Psi_\gamma(\bz)\big|_{\gammahat(\bz)}\,\nabla\gammahat(\bz),
  \label{eq:supp_chain}
\end{equation}
so $\bv$ differs from a gradient by a term along $\nabla\gammahat$; this is exactly what the level term in Eq.~\eqref{eq:supp_A_decomp} records.
Consequently \cref{prop:supp_two_sources} does \emph{not} say that $\bff^\star$ is of the form $\m\nabla\E$.
It says that the asymmetry of $\bff^\star$ splits into a part of that form, the scale term, which \ours{} represents exactly (\cref{sec:supp_class}), and a part that need not be, the level term.
By \cref{sec:supp_class}, where $\bff^\star$ is smooth and nonvanishing it is locally of the form $\m\nabla\E$ if and only if the level term vanishes in all pairs of directions orthogonal to $\bff^\star$; its component along $\bff^\star$ is absorbed by the gain.
The scale term is produced by the schedule: the same descent direction must be output with a magnitude $c(\gammahat)$ that varies across the input, and a landscape cannot encode a position-dependent magnitude without changing its directions too.
The level term vanishes when $\nabla\gammahat\parallel\bw$ or when $\gammahat$ is locally constant, and is small when $\partial_\gamma\nabla\Psi_\gamma$ is small; in the additive-noise setting of \cref{sec:supp_exact} the obstruction vanishes identically, since the field is exactly of the form $\m\nabla\E$.
Whether it is small for images is an empirical question that we do not measure directly; the generation results of \cref{sec:generation_quality} show that the fields representable with the scale term alone suffice to close most of the quality gap to implicit EqM.

\section{The Field Class of \ours{}}
\label{sec:supp_class}

\begin{proposition}[Fields of the form $m\nabla E$]
\label{prop:supp_class}
Let $\bff=m\nabla E$ with $m\in C^1(\R^D;\R_{>0})$ and $E\in C^2$.
Then
\begin{equation}
  \bA{\bff}=\tfrac12\big(\nabla E\,\nabla m^{\!\top}-\nabla m\,\nabla E^{\!\top}\big),
  \label{eq:supp_A_ceqm}
\end{equation}
which has rank at most two and satisfies $\bu^{\!\top}\bA{\bff}\bv=0$ for all $\bu,\bv\perp\bff(\bz)$.
Conversely, every field of the form $s\nabla\Psi$ with $s>0$ is of this form, with $m=s$ and $E=\Psi$.
The representation is not unique: for any $C^2$ function $h$ with $h'>0$, the pairs $(E,m)$ and $(h\circ E,\ m/h'(E))$ define the same field.
Finally, $\bff(\bz)=0$ if and only if $\nabla E(\bz)=0$; a strictly positive $m$ can make $\norm{\bff}$ arbitrarily small without changing the critical points of $E$.
\end{proposition}
\begin{proof}
$\bJ{\bff}=\nabla E\nabla m^{\top}+m\nabla^2E$, and the second term is symmetric, which gives Eq.~\eqref{eq:supp_A_ceqm}, \ie, Eq.~\eqref{eq:skew}.
The rank bound and the orthogonality follow as in \cref{prop:supp_two_sources}(b),(d), since $\nabla E\parallel\bff$. For the gauge freedom, $\nabla(h\circ E)=h'(E)\nabla E$, so $(m/h'(E))\nabla(h\circ E)=m\nabla E$. The last claim is $m>0$.
\end{proof}

Comparing Eq.~\eqref{eq:supp_A_ceqm} with Eq.~\eqref{eq:supp_A_decomp}, \ours{} produces asymmetry of exactly the form of the scale term and of no other form; in particular, it cannot produce asymmetry in pairs of directions orthogonal to its own field.
This is the precise content of the three rows of \cref{tab:classes}: \eqme{} has $\bA{\bff}\equiv0$, \ours{} has the rank-two structure of Eq.~\eqref{eq:supp_A_ceqm}, and implicit EqM is unrestricted.
Let $\bff$ be smooth and nonvanishing on an open set $U$.
Then $\bff$ admits, around every point of $U$, a local representation $m\nabla E$ with $m>0$ if and only if $\bu^{\top}\bA{\bff}\bv=0$ whenever $\bu,\bv\perp\bff$, throughout $U$: by Frobenius' theorem this is the integrability condition $\alpha\wedge d\alpha=0$ for the one-form $\alpha=\bff\cdot d\bz$, \ie, the condition that $\bff$ admits an integrating factor. (Smoothness matters: a $C^1$ field satisfying the condition need not have a $C^2$ potential.) For $D=2$ the condition holds automatically, so every nonvanishing planar field is locally of this form; the two-dimensional examples of \cref{fig:toy_example} therefore exhibit the restriction of \eqme{}, not that of \ours{}, which is a genuine restriction only for $D\ge3$ and, through zeros and global topology, in general. The gauge freedom means that the loss determines only the product $m\nabla E$; the split between magnitude in $\m$ and magnitude in $\nabla\E$ that training finds is a property of the parameterization, initialization, and optimizer, not of the objective. The behavior reported in \cref{sec:stop}, where $\m$ decays by two orders of magnitude while $\norm{\nabla\E}/\sqrt{d}$ does not decay, is therefore an empirical finding, consistent with the last statement of \cref{prop:supp_class}: the sampler slows down through $\m$, not because it reaches a critical point of $\E$.

\section{An Exact Case: Additive Gaussian Noise with Unknown Scale}
\label{sec:supp_exact}

\begin{proposition}[Exact factorization]
\label{prop:supp_exact}
Let $\x\sim\pdata$ with finite second moment, $\eps\sim\mathcal{N}(\bm{0},\bI)$, and $\sigma\sim\nu$ independent, with $\nu$ supported in $[\sigma_{\min},\infty)$ for some $\sigma_{\min}>0$ and $\int\sigma^2\nu(d\sigma)<\infty$.
Let $\bz=\x+\sigma\eps$, let $p_\sigma$ be the density of $\bz$ given $\sigma$, and define
\begin{equation}
  q(\bz)=\int p_\sigma(\bz)\,\nu(d\sigma),\qquad
  S(\bz)=\int\sigma^2p_\sigma(\bz)\,\nu(d\sigma).
  \label{eq:supp_qS}
\end{equation}
Then $q$ and $S$ are smooth and positive, and
\begin{equation}
\begin{aligned}
  \Ex[\bz-\x\mid\bz]&=m^\star(\bz)\,\nabla E^\star(\bz),\\
  E^\star(\bz)&=-\log S(\bz),\\
  m^\star(\bz)&=\frac{S(\bz)}{q(\bz)}=\Ex[\sigma^2\mid\bz],
\end{aligned}
  \label{eq:supp_exact}
\end{equation}
which is Eq.~\eqref{eq:exact_case}.
\end{proposition}
\begin{proof}
Given $\sigma$, Tweedie's formula~\cite{efron2011tweedie} gives $\Ex[\x\mid\bz,\sigma]=\bz+\sigma^2\nabla\log p_\sigma(\bz)$, so $\Ex[\bz-\x\mid\bz,\sigma]=-\sigma^2\nabla p_\sigma(\bz)/p_\sigma(\bz)$.
With the posterior $p(\sigma\mid\bz)=p_\sigma(\bz)\nu(\sigma)/q(\bz)$,
\begin{equation*}
\begin{aligned}
  \Ex[\bz-\x\mid\bz]
  &=-\frac{1}{q(\bz)}\int\sigma^2\nabla p_\sigma(\bz)\,\nu(d\sigma)
  =-\frac{\nabla S(\bz)}{q(\bz)}\\
  &=\frac{S(\bz)}{q(\bz)}\,\big(-\nabla\log S(\bz)\big)
  =m^\star(\bz)\,\nabla E^\star(\bz),
\end{aligned}
\end{equation*}
where differentiation under the integral is justified by $\sigma\ge\sigma_{\min}$ and the second-moment assumption, and $S/q=\int\sigma^2p(\sigma\mid\bz)\,d\sigma=\Ex[\sigma^2\mid\bz]$.
\end{proof}

The lower bound on $\sigma$ is a sufficient condition for $q$ and $S$ to be finite and smooth for every $\pdata$ with finite second moment; without it they can diverge (for $D=3$, $\pdata=\delta_{\bm 0}$, and $\sigma\sim\mathcal U(0,1)$, $q(\bm 0)\propto\int_0^1\sigma^{-3}d\sigma=\infty$).
The factorization is exact for every $D$ and every data distribution, without any concentration argument, and the obstruction of \cref{sec:supp_class} vanishes identically in this setting. It is in the sense that Eq.~\eqref{eq:factorized_field} has ``exactly the form of the optimal blind denoiser'' (\cref{sec:method}): the energy is a log-density-like landscape, and the gain is the posterior mean $\bar\sigma^2(\bz)=\Ex[\sigma^2\mid\bz]$ of the noise variance, \ie, a noise-level estimate made from the corrupted input alone. The setting differs from EqM in the corruption ($\x+\sigma\eps$ instead of $\gamma\x+(1-\gamma)\eps$) and in the target (the residual $\bz-\x$ instead of $c(\gamma)(\eps-\x)$). \Cref{sec:supp_two_terms} shows that the shrinkage factor $\gamma$ is what introduces the second term in the EqM case; the additive-noise case is therefore a sufficient example of when one scalar is enough, not a universal guarantee for EqM.

\section{Energy Descent Along the Sampler}
\label{sec:supp_descent}

\begin{proposition}[Energy descent]
\label{prop:supp_descent}
Let $E\in C^1(\R^D)$ be bounded below by $E_{\inf}$, let $m:\R^D\to\R_{>0}$ be continuous, and let $\bff=m\nabla E$.
\begin{itemize}
\item[(i)] Along any solution of $\dot\x=-\bff(\x)$,
\begin{equation}
  \frac{d}{dt}E(\x(t))=-m(\x)\norm{\nabla E(\x)}^2\le0,
  \label{eq:supp_lyapunov}
\end{equation}
with equality if and only if $\nabla E(\x)=0$; the stationary points of the dynamics are exactly the critical points of $E$.
\item[(ii)] Suppose in addition that $\nabla E$ is $L$-Lipschitz on a convex set that contains the iterates and that $0<m_{\min}\le m\le m_{\max}$ there.
Then $\x_{k+1}=\x_k-\eta\bff(\x_k)$ with $0<\eta\le1/(Lm_{\max})$ satisfies
\begin{equation}
  E(\x_{k+1})\le E(\x_k)-\frac{\eta\,m(\x_k)}{2}\norm{\nabla E(\x_k)}^2
  \label{eq:supp_step}
\end{equation}
for every $k$, and
\begin{equation}
  \min_{0\le k<K}\norm{\nabla E(\x_k)}^2\le\frac{2\,(E(\x_0)-E_{\inf})}{\eta\,m_{\min}\,K}.
  \label{eq:supp_rate}
\end{equation}
\end{itemize}
\end{proposition}
\begin{proof}
(i) is the chain rule together with $m>0$.
(ii) The descent lemma $E(\bm{y})\le E(\x)+\nabla E(\x)\cdot(\bm{y}-\x)+\tfrac L2\norm{\bm{y}-\x}^2$ with $\bm{y}-\x=-\eta m_k\nabla E(\x_k)$ and $m_k=m(\x_k)$ gives
$E(\x_{k+1})\le E(\x_k)-\eta m_k\big(1-\tfrac{L\eta m_k}{2}\big)\norm{\nabla E(\x_k)}^2$, and $L\eta m_k/2\le\tfrac12$.
Summing Eq.~\eqref{eq:supp_step} over $k<K$ and telescoping, $\tfrac\eta2m_{\min}\sum_{k<K}\norm{\nabla E(\x_k)}^2\le E(\x_0)-E_{\inf}$, and the sum is at least $K$ times its smallest term.
\end{proof}

\paragraph{Remarks}
(a) The softplus head of Eq.~\eqref{eq:gain} guarantees $m>0$ but no uniform bounds; on any compact region, which the iterates of a bounded trajectory stay in, the bounds in (ii) hold by continuity, with constants that depend on the model.
(b) The guarantee is for the plain gradient-descent sampler used on ImageNet (\cref{sec:setup}).
Heun's method, used on CIFAR, and Nesterov acceleration evaluate $\bff$ at intermediate or look-ahead points and are not monotone descent methods; even for a quadratic energy the energy can increase on individual steps.
(c) The guarantee is for the deterministic sampler.
A Langevin-type extension on the Riemannian manifold of \cref{sec:supp_geometry} would require an additional drift term proportional to $\nabla m$ to leave the intended distribution invariant~\cite{girolami2011riemann,li2016psgld}.
(d) The statement concerns the \emph{learned} energy $\E$; it does not make $\E$ a calibrated density or score, which is why the energy-based applications of \cref{sec:composition} and the OOD experiments are evaluated empirically.

\begin{figure*}[t]
  \centering
  \includegraphics[width=0.95\textwidth]{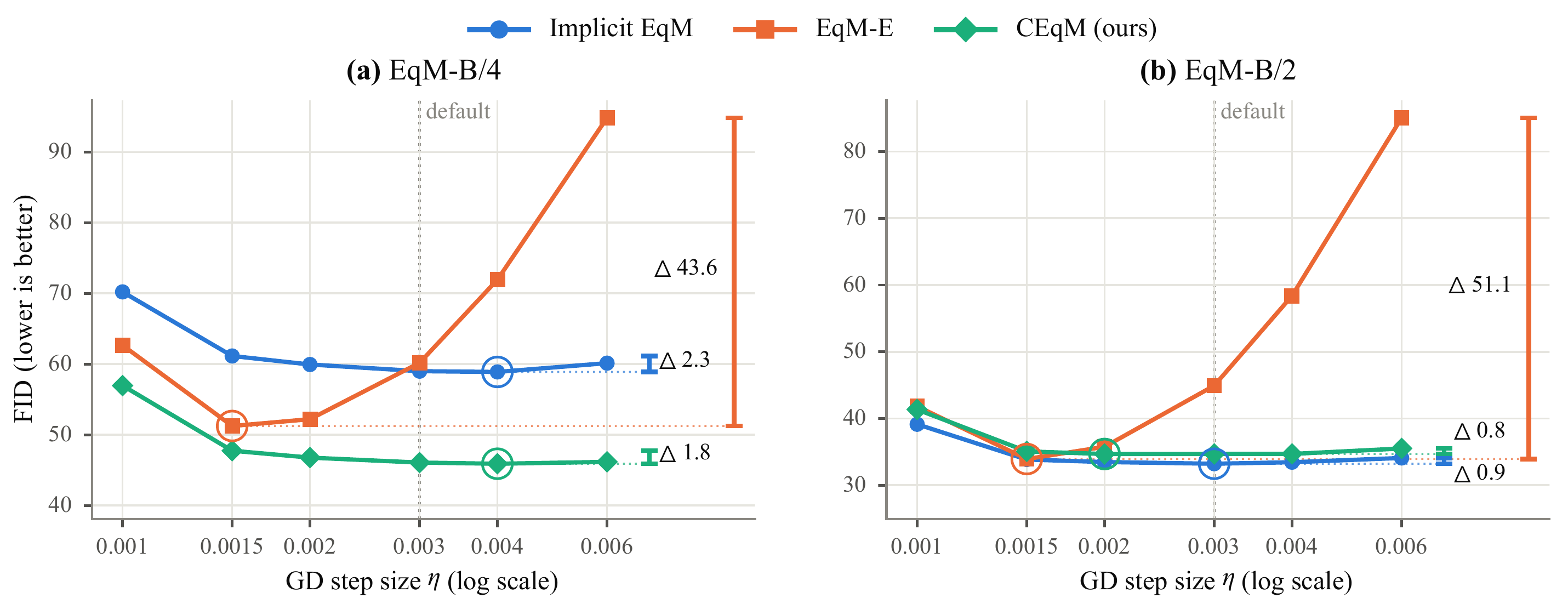}
  \caption{\textbf{Step-size robustness.} FID-50K against the gradient-descent step size $\eta$ (250 steps, no guidance) for \textbf{(a)}~EqM-B/4 and \textbf{(b)}~EqM-B/2 after 400k iterations; implicit EqM-B/2 is the official checkpoint. The dotted line marks EqM's default~($\eta = 0.003$), used throughout the main text, and rings mark each model's best $\eta$. Bars give each model's FID range $\Delta$ over $\eta \in [0.0015, 0.006]$.}
  \label{fig:supp_eta}
\end{figure*}

\section{Riemannian-Geometric Interpretation}
\label{sec:supp_geometry}

Let $g_{\x}(\bu,\bv)=\bu^{\top}\bm{G}(\x)\bv$ be a Riemannian metric on $\R^D$ with $\bm{G}(\x)$ symmetric positive definite.
The Riemannian gradient of $E$ is defined by $g_{\x}(\grad_gE,\bv)=\nabla E\cdot\bv$ for all $\bv$, hence $\grad_gE=\bm{G}^{-1}\nabla E$~\cite{lee2013smooth}; $\bm{G}=\bI$, $\nabla^2E$, and the Fisher matrix give the Euclidean gradient, the Newton direction, and the natural gradient~\cite{amari1998natural}.
A metric $\bm{G}=\varphi(\x)\bI$ with $\varphi>0$ is \emph{conformally flat}: it rescales lengths at each point by the same factor in all directions and preserves angles.
With $\varphi=m^{-1}$,
\begin{equation}
  \grad_gE=m\nabla E=\bff ,
  \label{eq:supp_riem}
\end{equation}
so the field of Eq.~\eqref{eq:factorized_field} is the Riemannian gradient of $\E$ under the conformal metric $\m^{-1}\bI$, the gain $\m$ is the reciprocal of the conformal factor, and the sampler of \cref{sec:method_train_and_sample} is Riemannian gradient descent on $\E$.
(``Conformal'' refers to the metric; it is unrelated to \emph{conformal vector fields}, whose flows preserve a metric up to scale.)

For a \emph{fixed} energy, the continuous-time dynamics $\dot\x=-m\nabla E$ can be read in three equivalent ways: Riemannian gradient flow under $m^{-1}\bI$; Euclidean gradient flow with a position-dependent rate $m(\x)$; or a time reparameterization $d\tau=m(\x(t))\,dt$ of the Euclidean gradient flow $d\x/d\tau=-\nabla E$, which shows that the trajectories are the same curves traversed at different speeds. This is the sense in which the gain is a learned, per-sample step size (\cref{sec:method_train_and_sample}). Two qualifications apply: the discrete iterates of the two flows with a fixed $\eta$ are not the same points, and during joint training $\E$ itself changes, so \ours{} is not equivalent to \eqme{} with a modified step-size schedule. Geometrically, $\norm{\delta\x}_g=\norm{\delta\x}/\sqrt{m(\x)}$: where $m$ is small, near the data, distances are stretched and a unit Riemannian step is a short Euclidean step, which is the brake of \cref{fig:teaser}; where $m$ is large, distances are compressed.

\section{Step-Size Robustness} \label{sec:supp_eta}
We evaluate the sensitivity of each variant to the gradient descent sampling step size $\eta$. \Cref{fig:supp_eta} reports the FID across $\eta \in \{0.001, 0.0015, 0.002, 0.003, 0.004, 0.006\}$, with each variant's sensitivity measured by the range between its best and worst FID over $\eta \in [0.0015, 0.006]$. \ours{}'s FID varies by only 1.8 at EqM-B/4 and 0.8 at EqM-B/2, comparable to implicit EqM (2.3 and 0.9, respectively). In contrast, \eqme{} exhibits greater sensitivity, with its range 43.6 and 51.1, respectively. \eqme{} achieves its best performance at the smallest step size in this range, but FID degrades sharply as $\eta$ increases. This behavior is consistent with the analysis in \cref{sec:stop}: because its field does not slow down at the data, larger steps can move samples beyond the real data distribution. Although tuning $\eta$ recovers much of \eqme{}'s performance (FID of 51.24 at EqM-B/4 and 33.93 at EqM-B/2, compared with 45.90 and 34.68 for \ours{}, respectively), this improvement is confined to a narrow range of step sizes.

\section{Implementation Details} \label{sec:supp_impl}
\Cref{tab:supp_hparams} lists the training and sampling settings. Within each column, the three variants share the same settings.

\begin{table*}[t]
  \centering
  \small
  \caption{Composition success rate (\%) over guidance weights and step sizes. Share of 2{,}500 samples (50 random class pairs $\times$ 50 samples) whose ResNet-50 top-5 predictions contain both classes. We report $w = 2$ at $\eta = 0.003$ in the main text.}
  \label{tab:supp_compose}
  \setlength{\tabcolsep}{5pt}
  \begin{tabular*}{\tabstretch\textwidth}{@{\extracolsep{\fill}}lc ccccc ccccc@{}}
    \toprule
    & & \multicolumn{5}{c}{EqM-B/4, step size $\eta$} & \multicolumn{5}{c}{EqM-B/2, step size $\eta$} \\
    \cmidrule(lr){3-7}\cmidrule(l){8-12}
    Model & $w$ & 0.0008 & 0.001 & 0.0015 & 0.002 & 0.003 & 0.0008 & 0.001 & 0.0015 & 0.002 & 0.003 \\
    \midrule
    Implicit EqM & 1 & 0.4 & 2.8 & \textbf{3.1} & 3.0 & 2.9 & 2.4 & 7.2 & 8.2 & 8.4 & \textbf{8.6} \\
    & 2 & 4.6 & 12.0 & \textbf{12.8} & 12.7 & 12.5 & 9.6 & 26.8 & \textbf{29.7} & 29.4 & 29.4 \\
    & 3 & 8.9 & 18.6 & \textbf{19.9} & 18.4 & 14.9 & 17.6 & 37.3 & \textbf{40.6} & 39.6 & 32.2 \\
    \midrule
    \eqme{} & 1 & 1.4 & 4.6 & \textbf{4.9} & 4.0 & 2.2 & 2.6 & \textbf{7.8} & 7.0 & 5.7 & 3.4 \\
    & 2 & 4.5 & \textbf{12.0} & 9.5 & 5.1 & 1.8 & 6.9 & \textbf{20.1} & 13.0 & 6.0 & 1.4 \\
    & 3 & 5.8 & \textbf{10.0} & 5.5 & 2.6 & 0.4 & 8.5 & \textbf{14.8} & 6.6 & 2.3 & 0.6 \\
    \midrule
    \textbf{\ours{}~(ours)} & 1 & 1.0 & 4.8 & 6.2 & \textbf{6.8} & 6.6 & 1.8 & 7.0 & 8.4 & 8.8 & \textbf{8.9} \\
    & 2 & 6.3 & 21.3 & 24.6 & \textbf{25.0} & 24.2 & 8.9 & 27.5 & 31.7 & \textbf{32.1} & 31.8 \\
    & 3 & 8.1 & 24.4 & \textbf{28.2} & 27.2 & 26.0 & 9.5 & 25.6 & \textbf{28.8} & 28.6 & 27.0 \\
    \bottomrule
  \end{tabular*}
\end{table*}

\begin{table*}[t]
\centering
\normalsize
\setlength{\tabcolsep}{5pt}
\renewcommand{\arraystretch}{1.3}
\caption{Training and sampling settings. All settings are shared across the three variants. CIFAR dataset settings follow~\cite{lipman2023flow}, except for the gradient schedule, which follows~\cite{wang2025eqm}.}
\label{tab:supp_hparams}
\begin{tabular}{@{}lcc@{}}
\toprule
& ImageNet-1K~$256\times256$ & CIFAR-10/100 \\
\midrule
\multicolumn{3}{@{}l}{\textit{Training}} \\
Architecture & EqM-B/4, EqM-B/2 & U-Net of~\cite{lipman2023flow} \\
Conditioning & label dropout 0.1 & unconditional \\
Optimizer & \multicolumn{2}{c}{AdamW} \\
Learning rate & \multicolumn{2}{c}{$10^{-4}$} \\
Betas $(\beta_1, \beta_2)$ & $(0.9, 0.999)$ & $(0.9, 0.95)$ \\
Weight decay & 0.0 & 0.01 \\
Batch size & \multicolumn{2}{c}{256} \\
EMA decay & 0.9999 & 0.999 \\
Iterations & 400k~($\approx$80 epochs) & 200k~($\approx$1000 epochs) \\
Schedule $c(\gamma)$ & truncated & piecewise \\
\quad $a$ & \multicolumn{2}{c}{0.8} \\
\quad $b$ & -- & 0.4 \\
\quad $\lambda$ & 4.0 & 2.0 \\
\midrule
\multicolumn{3}{@{}l}{\textit{Sampling}} \\
Sampler & gradient descent & Heun (2nd order) \\
Steps & 250 & 50 \\
Step size & $\eta = 0.003$ & $\Delta t = 0.02$ over $[0, 1]$ \\
\bottomrule
\end{tabular}
\end{table*}

\paragraph{Data}
For ImageNet, images are center-cropped to $256\times256$ as in \cite{dhariwal2021diffusion} and then encoded with the Stable Diffusion VAE into $32\times32\times4$ latents, scaled by 0.18215. We cache the latent representations of each image and its horizontal flip, and pick one of the two at random in each iteration. For the CIFAR datasets, we train directly in pixel space and similarly use horizontal flips as data augmentation.

\paragraph{Sampling and evaluation}
We evaluate all generation metrics using the ADM evaluation suite~\cite{dhariwal2021diffusion}. For each experiment, we generate 50{,}000 samples using class labels drawn uniformly at random. All models use the same sampling seed and generate samples from their respective EMA weights.

\paragraph{Compute}
All training runs were conducted on a single NVIDIA H200. On ImageNet, implicit trains at 12.9 iterations per second for the EqM-B/4 and B/2 backbones, and at 4.2 iterations per second for \eqme{} and \ours{}. The explicit variants need a second backward pass to differentiate the input gradient of the energy. Similarly, sampling with the explicit variants requires an additional backward pass per step.

\section{Compositional Generation Details} \label{sec:supp_compose}
\begin{figure*}[t]
  \centering
  \includegraphics[width=\textwidth]{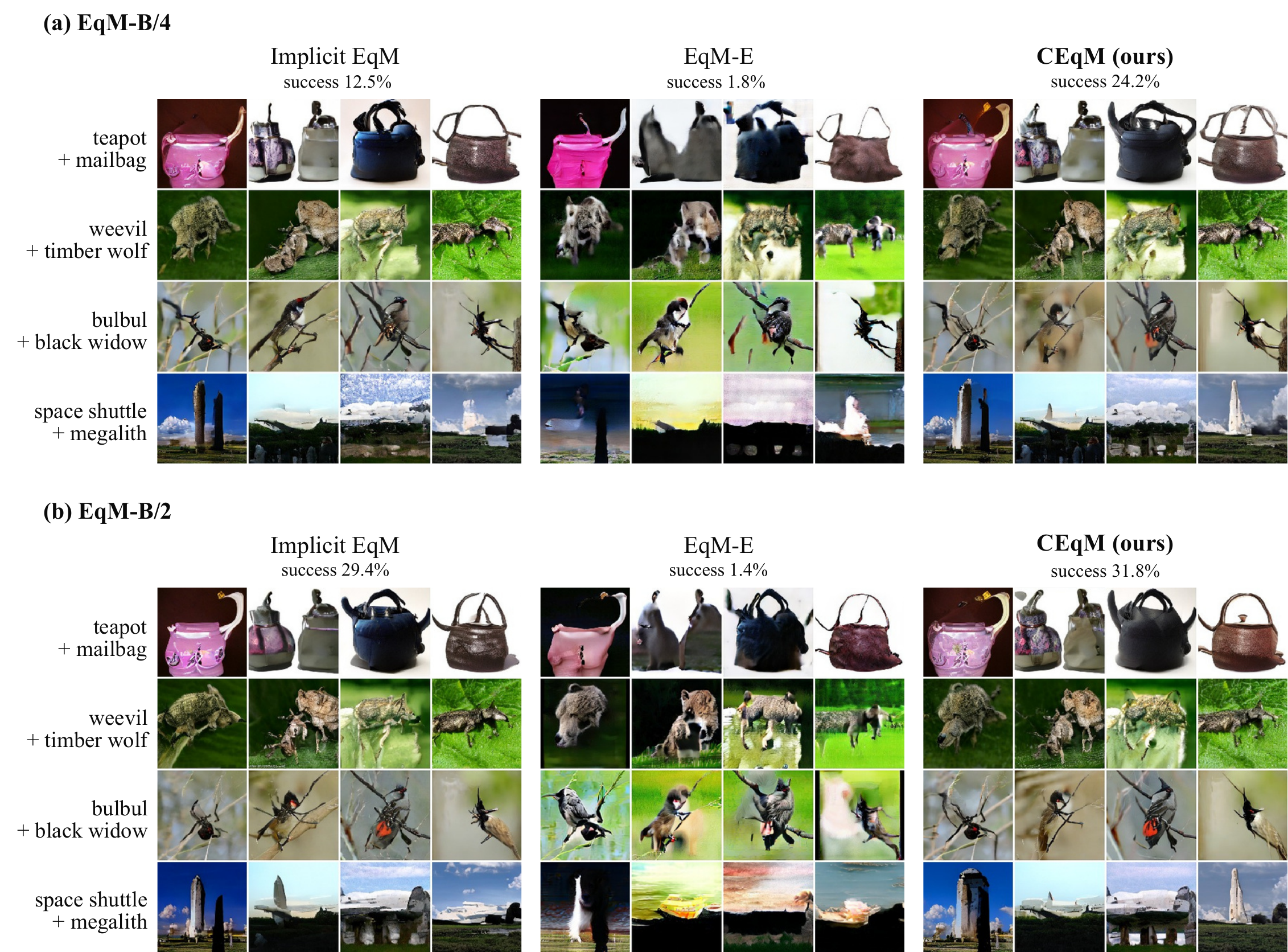}
  \caption{\textbf{Uncurated compositional samples.} The first four of the 50 class pairs (rows), every two classes drawn uniformly at random with a fixed seed, and the first four samples of each, at $w = 2$ and $\eta = 0.003$ (250 steps), with the same starting noise for every model. Implicit EqM mixes its fields; \eqme{} and \ours{} sum their energies~(Eq.~\ref{eq:compose_conjunction}). Written above each grid is the composition success rate as described in~\cref{sec:supp_compose}.}
  \label{fig:compose_grid}
\end{figure*}

We sample the conjunction of two ImageNet classes $y_1, y_2$ using \cref{eq:compose_conjunction}. We draw 50 random class pairs with a fixed seed shared by all models, and 50 samples per pair from the same starting noise for each model, yielding 2{,}500 samples per model and setting. We classify each sample by a pretrained ResNet-50 and count it as a success if both classes are among its top-5 predictions. \Cref{fig:compose_grid} shows uncurated examples: the first four of the 50 class pairs, each drawn uniformly at random with a fixed seed, and the first four samples of each, from the same starting noise for every model.

\clearpage
\begin{figure*}[t]
  \centering
  \includegraphics[width=\textwidth]{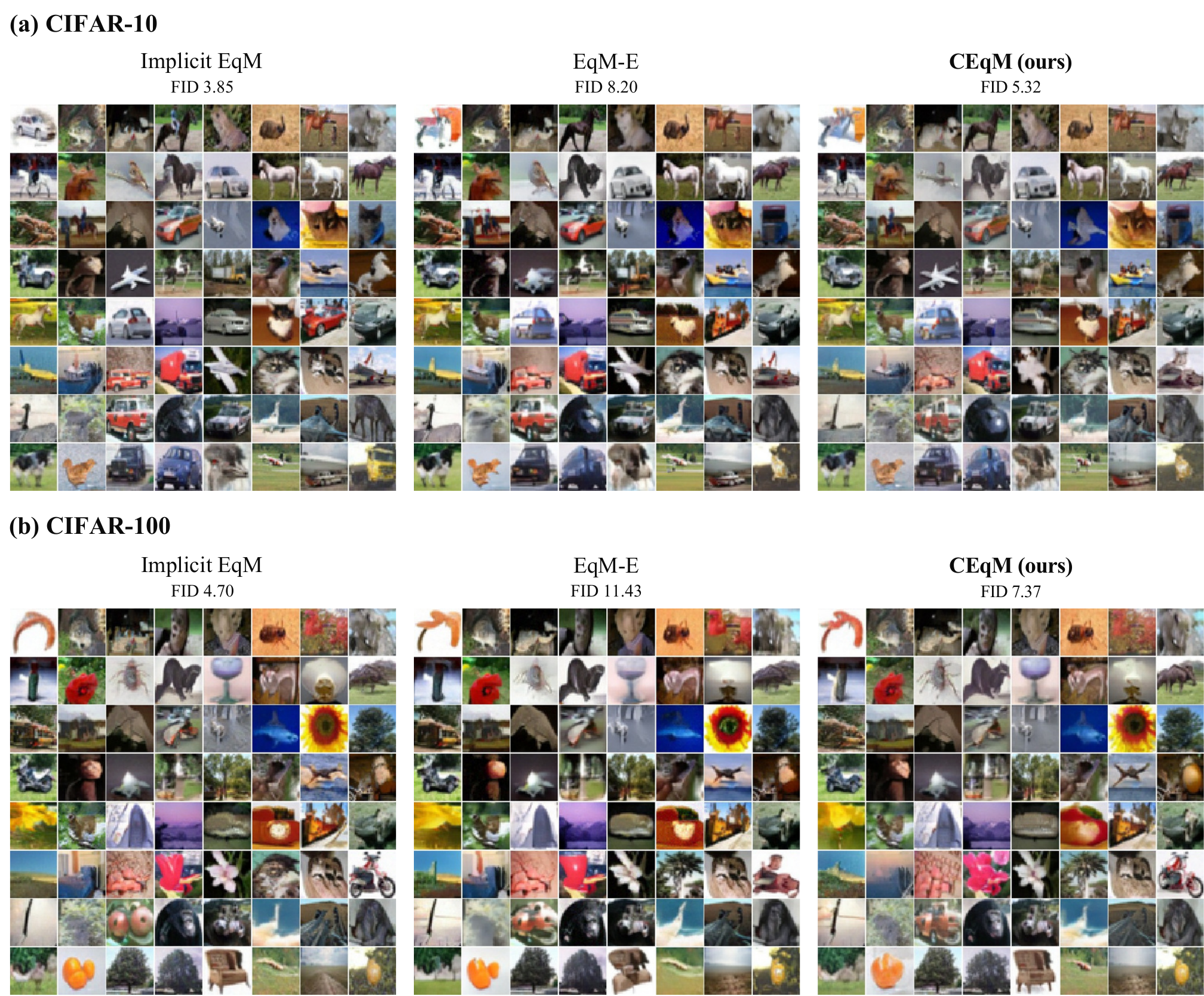}
  \caption{\textbf{Uncurated samples, CIFAR Datasets.} 64 randomly selected samples from each model's 50K-sample evaluation set.}
  \label{fig:supp_uncurated_cifar}
\end{figure*}

\clearpage
\begin{figure*}[t]
  \centering
  \includegraphics[width=\textwidth]{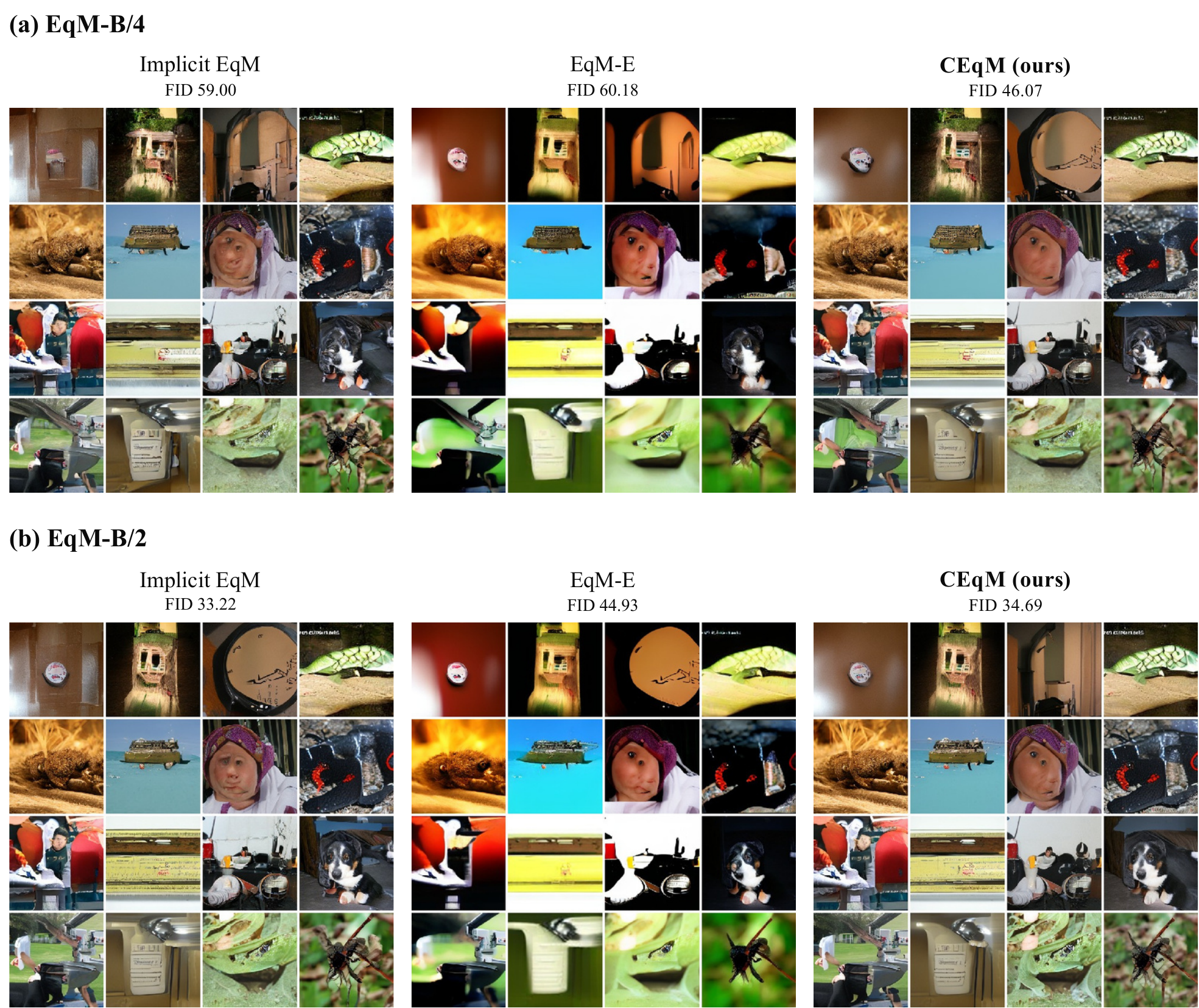}
  \caption{\textbf{Uncurated samples, ImageNet-1K 256$\times$256.} 16 randomly selected samples from each model's 50K-sample evaluation set. All models were trained according to the settings described in \cref{tab:supp_hparams}, except for implicit EqM-B/2, for which we use the official checkpoint.}
  \label{fig:supp_uncurated_imagenet}
\end{figure*}

\clearpage

\clearpage 
\bibliographystyle{IEEEtran}
\bibliography{references}

\end{document}